\documentclass[10pt,twocolumn,letterpaper]{article}

\usepackage[pagenumbers]{cvpr} 

\usepackage{lipsum}
\usepackage{array}
\usepackage{times}
\usepackage{epsfig}
\usepackage{graphicx}
\usepackage{amsmath,amssymb,amsthm}
\usepackage{algorithm,algorithmicx,algpseudocode}
\usepackage{bm,xspace}
\usepackage{comment}
\usepackage{multirow}
\usepackage{balance}
\usepackage{url}
\usepackage{booktabs}
\usepackage{etoolbox,siunitx}
\usepackage{calc}
\usepackage{pifont,hologo}
\usepackage{color}
\usepackage{adjustbox}
\usepackage{bbding}  %
\usepackage[normalem]{ulem}  %
\usepackage{contour}
\usepackage[table]{xcolor}
\usepackage{soul}%
\usepackage{tocloft} %
\usepackage{etoc} %
\PassOptionsToPackage{dvipsnames}{xcolor}
\usepackage[pagebackref,breaklinks,colorlinks,allcolors=cvprblue]{hyperref}

\usepackage{fancyhdr}

\definecolor{blue}{HTML}{004bb3}
\definecolor{red}{HTML}{cc1100}
\definecolor{orange}{HTML}{cc7700}
\definecolor{gray}{HTML}{efefef}
\definecolor{darkgreen}{HTML}{228B22}
\definecolor{darkgray}{HTML}{757575}

\definecolor{cite}{HTML}{3270b5}
\definecolor{link}{HTML}{b53532}
\definecolor{link}{HTML}{cc1100}
\definecolor{scratch}{HTML}{001219}
\definecolor{pretrain}{HTML}{0A9396}
\definecolor{sm}{HTML}{001219}
\definecolor{smpro}{HTML}{36B0A4}
\definecolor{smap50}{HTML}{1f77b4}

\definecolor{fe}{HTML}{ECEDFE}
\definecolor{qi}{HTML}{EAF2FA}
\definecolor{qr}{HTML}{A9EAF9}

\definecolor{m_pink}{RGB}{253, 244, 244}
\definecolor{m_blue}{RGB}{234, 242, 250}
\definecolor{m_gray}{RGB}{242, 242, 244}

\definecolor{ourscolor}{HTML}{DCFCF9}
\definecolor{offsetcolor}{HTML}{6424D6}

\definecolor{backbone}{RGB}{220, 234, 247}
\definecolor{saqs}{RGB}{234, 242, 250}
\definecolor{psqd}{RGB}{202, 238, 251}

\contourlength{0.8pt}

\newcommand{\figref}[1]{Fig.~\ref{#1}}
\newcommand{\tabref}[1]{Tab.~\ref{#1}}
\newcommand{\secref}[1]{Sec.~\ref{#1}}
\renewcommand{\eqref}[1]{Eq.~\ref{#1}}

\newcolumntype{x}[1]{>{\centering\arraybackslash}p{#1}}
\newcolumntype{y}[1]{>{\raggedright\arraybackslash}p{#1}}
\newcolumntype{z}[1]{>{\raggedleft\arraybackslash}p{#1}}
\newcommand{\tablestyle}[2]{\setlength{\tabcolsep}{#1}\renewcommand{\arraystretch}{#2}\centering\footnotesize}

\DeclareMathSymbol{@}{\mathord}{letters}{"3B}

\newcommand\mypara[1]{\vspace{0mm}\noindent\textbf{#1}}

\newcommand{\YesV}{\ding{51}}%
\newcommand{\NoX}{\ding{55}}%

\makeatletter
\DeclareRobustCommand\onedot{\futurelet\@let@token\@onedot}
\def\@onedot{\ifx\@let@token.\else.\null\fi\xspace}
 
\def\ie{\emph{i.e}\onedot} 
\def\cf{\emph{cf}\onedot} 
 \def\vs{\emph{vs}\onedot}

\newcommand*{\Rom}[1]{\expandafter\@slowromancap\romannumeral #1@}
\newcommand*{\rom}[1]{\expandafter\romannumeral #1}

\def\1{\bm{1}}

\def\rmI{{\mathbf{I}}}

\def\rmP{{\mathbf{P}}}

\def\rmT{{\mathbf{T}}}

\def\ve{{\bm{e}}}
\def\vf{{\bm{f}}}

\def\vh{{\bm{h}}}

\def\vk{{\bm{k}}}

\def\vp{{\bm{p}}}
\def\vq{{\bm{q}}}

\def\vs{{\bm{s}}}

\def\vv{{\bm{v}}}
\def\vw{{\bm{w}}}

\def\mP{{\bm{P}}}

\DeclareMathAlphabet{\mathsfit}{\encodingdefault}{\sfdefault}{m}{sl}
\SetMathAlphabet{\mathsfit}{bold}{\encodingdefault}{\sfdefault}{bx}{n}

\def\gE{{\mathcal{E}}}
\def\gF{{\mathcal{F}}}

\def\gL{{\mathcal{L}}}

\def\gT{{\mathcal{T}}}

\newcommand{\R}{\mathbb{R}}

\let\originalleft\left
\let\originalright\right
\renewcommand{\left}{\mathopen{}\mathclose\bgroup\originalleft}
\renewcommand{\right}{\aftergroup\egroup\originalright}

\definecolor{cvprblue}{rgb}{0.21,0.49,0.74}

\newcommand{\ours}{HAMMER\xspace}
\def\cont{{\texttt{[CONT]}}}

\newcommand{\authorskip}{\hspace{2.5mm}}
\newcommand{\institutionskip}{\hspace{5.0mm}}

\title{\raisebox{-0.2em}{\includegraphics[height=1.2em]{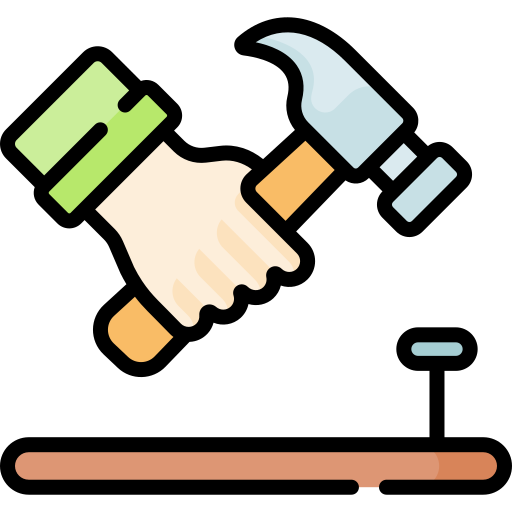}}~\ours: Harnessing MLLM via Cross-Modal Integration for Intention-Driven 3D Affordance Grounding}

\author{Lei Yao\textsuperscript{\mdseries1} \authorskip 
Yong Chen\textsuperscript{2} \authorskip
Yuejiao Su\textsuperscript{\mdseries1} \authorskip
Yi Wang\textsuperscript{\mdseries1*} \authorskip
Moyun Liu\textsuperscript{2} \authorskip
Lap-Pui Chau\textsuperscript{\mdseries1*} \\
\textsuperscript{1}The Hong Kong Polytechnic University \institutionskip
\textsuperscript{2}Huazhong University of Science and Technology \\ 
{\tt\small \url{https://rayyoh.github.io/Hammer/}}
}

\begin{document}
\maketitle
\renewcommand{\thefootnote}{\fnsymbol{footnote}}
\footnotetext[1]{Corresponding author.}
\begin{abstract}
    Humans commonly identify 3D object affordance through observed interactions in images or videos, and once formed, such knowledge can be generically generalized to novel objects. Inspired by this principle, we advocate for a novel framework that leverages emerging multimodal large language models (MLLMs) for interaction intention-driven 3D affordance grounding, namely \ours. Instead of generating explicit object attribute descriptions or relying on off-the-shelf 2D segmenters, we alternatively aggregate the interaction intention depicted in the image into a contact-aware embedding and guide the model to infer textual affordance labels, ensuring it thoroughly excavates object semantics and contextual cues. We further devise a hierarchical cross-modal integration mechanism to fully exploit the complementary information from the MLLM for 3D representation refinement and introduce a multi-granular geometry lifting module that infuses spatial characteristics into the extracted intention embedding, thus facilitating accurate 3D affordance localization. Extensive experiments on public datasets and our newly constructed corrupted benchmark demonstrate the superiority and robustness of \ours compared to existing approaches. All code and weights are publicly available.
\end{abstract}
\section{Introduction}
\label{sec:intro}
Object affordance refers to the properties indicating how it can be used or interacted with~\cite{gibson2014theory}. Intention-driven 3D affordance grounding~\cite{yang2023grounding} intends to anticipate potential actionable regions on point clouds based on corresponding interaction images, which attempts to mimic mankind's innate ability in perceiving object functionalities through observed demonstrations. This task naturally bridges the gap between visual comprehension and physical world interaction, facilitating the development of embodied agents for various applications, including robot dexterous manipulation~\cite{ma2024glover,zheng2025survey}, imitation learning~\cite{wu2025afforddp,bahl2023affordances}, and augmented reality~\cite{quesada2022proactive}.

\begin{figure}
    \centering
    \includegraphics[width=0.99\linewidth]{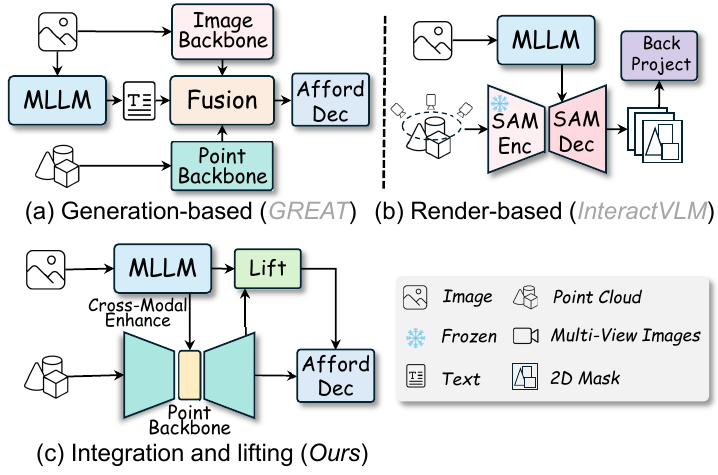}
    \caption{\textbf{Comparison of current architectures and \ours.} (a) GREAT~\cite{shao2024great} generates texts to assist the fusion process. (b) InteractVLM~\cite{dwivedi_interactvlm_2025} first produces 2D masks and back-projects them into 3D space. (c) \ours enhances point cloud features with cross-modal information from MLLMs and lifts the extracted intention embedding to 3D for accurate affordance localization.}
    \label{fig:teaser}
    \vspace{-1.0em}
\end{figure}

Precise 3D affordance grounding necessitates both comprehensive visual understanding and spatial cognition. The former focuses on deciphering object semantics, contextual cues, and interaction intentions, while the latter concerns the spatial and geometric interpretation of 3D structures. Although the provided image hints at crucial insights into human intention and interaction patterns, the target 3D object often exhibits significant variations in shape and scale. Thus, \textbf{\textit{refining object representation across modalities}} to ensure image features can complement 3D data presents a significant challenge. Furthermore, effectively \textbf{\textit{integrating interaction knowledge}} implied in the given image is another critical aspect for accurate affordance prediction. Current generation-based methods, like~\cite{shao2024great}, design separate branches for modality-specific perception and produce elaborate language descriptions of object properties and interaction contexts using pre-trained multimodal large language models (MLLMs)~\cite{bai2025qwen2,liu2023visual} as priors to promote the fusion process (\cf \figref{fig:teaser}a). However, it requires manually annotated templates and a two-stage training pipeline. Another line of work, such as~\cite{dwivedi_interactvlm_2025}, fine-tunes an off-the-shelf segmenter~\cite{lai2024lisa,kirillov2023segment} to yield 2D contact maps by rendering point clouds onto multi-view planes and back-projecting them into 3D space (\cf~\figref{fig:teaser}b). These existing frameworks either fail to fully exploit the powerful understanding capabilities of MLLMs or suffer from inevitable detail loss and error accumulation, resulting in suboptimal performance.

In this paper, we present \ours, a novel architecture for intention-driven 3D affordance grounding (\cf \figref{fig:teaser}c). Without resorting to intermediate texts or 2D masks, we process the image via an MLLM and aggregate the underlying intention into a contact-aware embedding. Given the MLLM's prominent multimodal perception capability, we adhere to auxiliary supervision to guide the model toward predicting textual affordance labels, ensuring it sufficiently seizes object semantics and contextual cues. Beyond using the MLLM solely for intention extraction, we further explore a hierarchical cross-modal integration mechanism to incorporate the knowledge encapsulated in its hidden states into point features, enhancing object representation alignment. This design is motivated by the observation that such hidden states inherently grasp task-relevant content, which helps alleviate the discrepancies arising from different modalities and promotes more coherent feature expression. Despite the obtained intention embedding already containing abundant 2D interaction clues, our intuition is that it lacks explicit spatial information, which is vital for precise affordance localization. To address this, we propose a multi-granular geometry lifting module that progressively infuses geometric characteristics into the embedding, thereby strengthening its 3D awareness. We then apply a decoder to jointly process the refined point features and the geometry-enhanced embedding for 3D affordance region prediction. Extensive experiments on standard datasets consistently show that \ours~outperforms state-of-the-art methods across multiple evaluation metrics and settings. Additionally, \ours exhibits remarkable resilience against various perturbations on our newly established corrupted benchmark with noise-infused point clouds.

In summary, our contributions are as follows:
\begin{itemize}
    \item We propose \ours, a new framework that extracts a contact-aware intention embedding and leverages knowledge from an MLLM to enrich 3D representations via a hierarchical cross-modal integration mechanism.
    \item We introduce a multi-granular geometry lifting module that injects different levels of spatial cues into the intention embedding for accurate 3D affordance localization.
    \item We evaluate \ours on standard datasets and a newly constructed corrupted benchmark, validating its consistent performance gains and robustness compared to existing approaches.
\end{itemize}

\section{Related Work}
\label{sec:related}

\subsection{Affordance Learning}
Affordance learning, which focuses on identifying potential actionable regions on objects, has attracted considerable interest across multiple research domains. UAD~\cite{tang2025uad} proposes an unsupervised distillation strategy that directly extracts task-conditioned affordance from vision foundation models like CLIP~\cite{radford2021learning} for 2D images. In parallel, GLOVER~\cite{ma2024glover} treats the output embeddings of large language models as specific prompts to generate affordance masks using SAM~\cite{kirillov2023segment}, extending the task to an open-vocabulary setting. Meanwhile, several studies have concentrated on affordance prediction from 3D point clouds. For instance, OpenAD~\cite{nguyen2023open} estimates affordance regions by calculating the similarity between linguistic descriptions and latent point features, whereas OVAD~\cite{van2024open} introduces a distillation mechanism to transfer knowledge from a teacher model pre-trained on large-scale 3D datasets. Moreover, LASO~\cite{li2024laso} investigates the fusion of linguistic and geometric features and designs a learnable query-driven decoder. GEAL~\cite{lu2024geal} utilizes 3D Gaussian Splatting (3DGS)~\cite{kerbl20233d} for image rendering to incorporate 2D perceptual cues and enhances representation learning by a cross-modal consistency alignment module, a paradigm similar to that of GaussianCross~\cite{yao2025gaussiancross}. SceneFun3D~\cite{delitzas2024scenefun3d} further extends the task to scene-level scenarios~\cite{yao2024sgiformer}, achieving affordance understanding in complex environments.

\subsection{Intention-Driven 3D Affordance Grounding}
Departing from language-prompted affordance prediction, IAGNet~\cite{yang2023grounding} proposes grounding 3D affordance from intentions depicted in interaction images and constructs the first benchmark, PIAD~\cite{yang2023grounding}. Subsequent work~\cite{shao2024great} scales it with additional affordance categories and 3D instances for more comprehensive evaluation. In order to improve generalization, MIFAG~\cite{gao2024learning} extracts invariant affordance knowledge from multiple reference images and maintains a dictionary for adaptive feature fusion. Beyond object affordance estimation, LEMON~\cite{yang2024lemon} synergistically models human contact and spatial relations by tailored interaction intention excavation and geometric correlation modules, enabling broader functional understanding. Further extending the scope, EgoChoir~\cite{yang2025egochoir} adapts this paradigm to egocentric views by incorporating head motion information. To reduce dependency on real images, ComA~\cite{kim2024beyond} and DAViD~\cite{kim2025david} leverage in-the-wild diffusion models to synthesize realistic human-object interaction samples, which are subsequently used for 3D affordance estimation. While our method is also built upon the task of grounding 3D affordance from interaction images, it distinctively focuses on harnessing the rich multimodal understanding of MLLMs to extract and leverage the nuanced information embedded in such images, setting it apart from prior works.

\subsection{MLLMs for 3D Affordance}
Benefit from large-scale pre-training, MLLMs have shown great potential across diverse tasks. As a pioneering exploration, PAVLM~\cite{liu2024pavlm} directly adds point features to instruction embeddings and fine-tunes LLaMa~\cite{touvron2023llama} for affordance prediction. 3D-ADLLM~\cite{chu20253d}, however, pre-trains the model using a referring object part segmentation task to acquire general knowledge, then fine-tunes it with LoRA~\cite{hu2022lora} for instruction-based reasoning affordance. In contrast, SeqAfford~\cite{yu2024seqafford} uses the intrinsic object understanding capability of ShapeLLM~\cite{qi2024shapellm} to address sequential affordance segmentation. For intention-driven 3D affordance grounding, GREAT~\cite{shao2024great} designs a chain-of-thought prompting strategy to extract object attributes and interaction descriptions from an MLLM, and proposes a multi-branch fusion strategy that integrates image, text, and point features. Although effective, it underutilizes the MLLM's inherent 2D comprehension capacity, relying instead on a separate image encoder. InteractVLM~\cite{dwivedi_interactvlm_2025} renders point clouds onto multi-view images, infers 2D affordance masks via LISA~\cite{lai2024lisa}, and lifts them to 3D by back-projection, though this may lead to geometric inconsistencies due to incomplete shape coverage. In comparison, our \ours fully utilizes the powerful reasoning and understanding of MLLMs by introducing a hierarchical cross-modal integration mechanism and a multi-granular geometry lifting module, enabling more effective extraction of image-based interaction cues and yielding more accurate 3D affordance predictions.
\section{Methodology}
\label{sec:method}

\begin{figure*}
    \centering
    \includegraphics[width=0.99\linewidth]{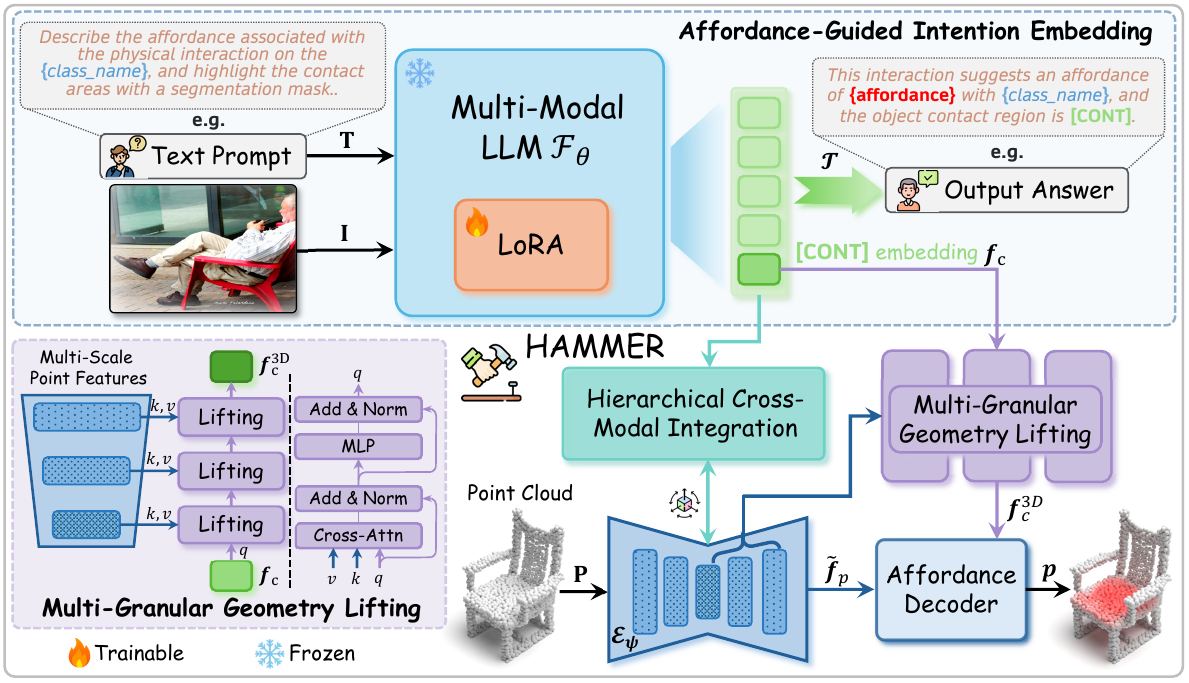}
    \caption{\textbf{Overview of our \ours.} Given a 3D point cloud $\rmP$ and its corresponding interaction image $\rmI$, our framework first processes $\rmI$ through a pre-trained MLLM $\gF_{\theta}$ to extract an affordance-guided intention embedding $\vf_c$ (\secref{subsec:intention-embedding}). This embedding is then used to enhance point cloud features via a hierarchical cross-modal integration mechanism (\secref{subsec:cross-modal-feature-enhancement}). To imbue $\vf_c$ with 3D spatial awareness, we apply a multi-granular geometry lifting module that incorporates multi-scale geometric cues (\secref{subsec:geometry-lifting}). Finally, the refined point features $\tilde{\vf}_p$ and the 3D-aware intention embedding $\vf_c^{3D}$ are decoded to produce the final affordance map $\vp$ (\secref{subsec:affordance-decoding}).}
    \label{fig:overview}
\end{figure*}

\subsection{Overview}
Given a point cloud $\rmP \in \R^{N \times 3}$ and its paired interaction image $\rmI \in \R^{H \times W \times 3}$, our goal is to ground a 3D affordance map $\vp = \{\vp_i\}_{i=0}^{N}$, where $\vp_i \in [0,1]$, $N$ is point number, $H$ and $W$ are the height and width of the image, respectively. We first process $\rmI$ using an MLLM to extract a contact-aware embedding $\vf_c \in \R^d$, which comprehensively excavates the hinted interaction intention (\secref{subsec:intention-embedding}), where $d$ is the feature dimension. To improve interaction understanding, we guide the MLLM to generate textual affordance labels as an auxiliary task. We then enhance the 3D point features via a hierarchical cross-modal integration mechanism using hidden states of the MLLM (\secref{subsec:cross-modal-feature-enhancement}). To alleviate geometric deficiency of the 2D-derived embedding $\vf_c$, we propose a multi-granular geometry lifting module (\secref{subsec:geometry-lifting}) that leverages multi-level spatial features to produce a 3D-aware embedding $\vf_c^{3D} \in \R^d$. Finally, we decode the enhanced point features $\tilde{\vf}_p \in \R^{N \times d}$ and lifted embedding $\vf_c^{3D}$ into the affordance map $\vp$ (\secref{subsec:affordance-decoding}). The overall architecture of \ours is illustrated in \figref{fig:overview}.

\subsection{Affordance-Guided Intention Embedding}
\label{subsec:intention-embedding}
Inspired by the success of MLLM in visual tasks~\cite{lai2024lisa,ma2024glover}, we employ a pre-trained MLLM $\gF_{\theta}$ to interpret the input image $\rmI$ and extract a contact-aware intention embedding $\vf_c$, where $\theta$ denotes the trainable parameters. Concretely, we extend the model's vocabulary by introducing a special token $\cont$ to aggregate interaction-related information. Since images often contain visually distracting elements that may impede learning, we design an object-centric prompting strategy that comprises object categorical labels in the text prompt $\rmT$, guiding the model to focus on relevant object semantics and contextual cues. An example of our prompting template is shown in~\figref{fig:overview}, where $\texttt{class\_name}$ is the given class prior and $\texttt{affordance}$ is the inferred affordance label. Without additional pre-processing, we feed the image-text pair $(\rmI, \rmT)$ into $\gF_{\theta}$ and extract the last-layer hidden states $\vh \in \R^{L \times d_h}$ for all tokens, where $L$ is the sequence length and $d_h$ is the hidden dimension. The intention embedding $\vf_c$ is obtained by projecting $\cont$'s hidden state $\vh_{\cont} \in \R^{d_h}$ via an MLP head $\psi_c$, \ie, $\vf_c = \psi_c(\vh_{\cont})$. Simultaneously, we request the model to generate an associated response $\gT$, incorporating the textual affordance label prediction $\gT_{\texttt{aff}}$ as an auxiliary task to enhance the model's awareness of task-related details. With available ground-truth labels and descriptions, we supervise the process with the standard language modeling loss $\gL_{txt}$~\cite{bai2025qwen2}. This technique effectively consolidates interaction-related information into $\vf_c$, supporting subsequent operations.

\subsection{Hierarchical Cross-Modal Integration}
\label{subsec:cross-modal-feature-enhancement}
Existing methods~\cite{yang2023grounding,lu2024geal,shao2024great} typically extract point features using 3D backbones, but they often lack sufficient semantic and interaction information, resulting in suboptimal performance. To overcome this limitation, we propose a hierarchical cross-modal integration mechanism that enhances point encoding by leveraging the hidden states $\vh$ of MLLM $\gF_{\theta}$, which possesses rich visual understanding and world knowledge. Adopting the standard encoder-decoder paradigm for point cloud processing, we introduce a two-stage integration strategy to fuse cross-modal information, thereby improving cross-model representation alignment.

The point cloud $\rmP$ is first processed by an encoder $\gE_{\psi}^{enc}$ to produce bottleneck features $\vf_p^{enc}$, while the hidden states $\vh$ are projected into a latent space via a shallow MLP $\phi_h$:
\begin{equation}
    \vf_p^{enc} = \gE_{\psi}^{enc}(\rmP) \in \R^{N_R \times d}, \quad \vf_h = \phi_h(\vh) \in \R^{L \times d},
\end{equation}
where $N_R$ indicates down-sampling point size and $d$ is the feature dimension. Here, $\vf_p^{enc}$ provides a compact global representation of the object geometry, and $\vf_h$ offers contextual cues regarding the interaction. To fully combine these complementary sources, we enhance $\vf_p^{enc}$ by incorporating $\vf_h$ via an attention-based mechanism. By treating $\vf_p^{enc}$ as the query and $\vf_h$ as the key and value, the cross-attention operation~\cite{vaswani2017attention} allows each point to selectively attend to relevant interaction cues, leading to more discriminative point representation:
\begin{equation}
    \tilde{\vf}_p^{enc} = \texttt{CrossAttn}(\vf_p^{enc}, \vf_h, \vf_h) \in \R^{N_R \times d}.
\end{equation}
The refined $\tilde{\vf}_p^{enc}$ is then passed to a decoder $\gE_{\psi}^{dec}$ to iteratively recover the full-resolution feature map $\vf_p^{(0)}$ and extract multi-scale features $\{\vf_p^{(i)}\}_{i=1}^R$ from intermediate layers:
\begin{equation}
    \{\vf_p^{(i)}\}_{i=1}^R, \vf_p^{(0)} = \gE_{\psi}^{dec}(\tilde{\vf}_p^{enc}),
\end{equation}
where $\vf_p^{(i)} \in \R^{N_i \times d}$ is the feature at layer $i$ with point count $N_i$ and $R$ is the number of layers. $\vf_p^{(0)} \in \R^{N \times d}$ is the output at the original point resolution $N$. This upsampling process helps restore fine spatial details, and the intermediate features $\{\vf_p^{(i)}\}_{i=1}^R$ capture geometric cues at multiple scales. We further refine $\vf_p^{(0)}$ to incorporate object-level semantics from $\vf_h$ for a more holistic representation. Since objects usually occupy a limited portion of the image, we use a gating mechanism to adaptively weight the tokens in $\vf_h$, yielding a global descriptor:
\begin{equation}
    \vs = \frac{\exp(\vf_h \cdot \vw_s)}{\sum_{m=1}^{L} \exp(\vf_{h,m} \cdot \vw_s)}, \quad \vf_h^g = \sum_{m=1}^{L} s_m \cdot \vf_{h,m},
\end{equation}
where $\vw_s \in \R^{d \times 1}$ is a learnable weight vector. The global descriptor $\vf_h^g$ is then duplicated to align with the spatial dimension of $\vf_p^{(0)}$:
\begin{equation}
    \vf_h^{g*} = \texttt{Duplicate}(\vf_h^g, N).
\end{equation}
Finally, we concatenate $\vf_p^{(0)}$ with $\vf_h^{g*}$ along the feature dimension and apply an MLP $\phi_f$ to achieve the final enhancement:
\begin{equation}
    \tilde{\vf}_p = \phi_f \left (\left [\vf_p^{(0)} || \vf_h^g \right ]\right ),
\end{equation}
where $\left [\cdot || \cdot \right ]$ means vector concatenation. The proposed hierarchical integration strategy effectively fuses multimodal cues at both global and local levels. The initial enhancement at the bottleneck stage enables point features to absorb rich contextual information, while the subsequent feature-level refinement promotes a comprehensive understanding of object semantics.

\subsection{Multi-Granular Geometry Lifting}
\label{subsec:geometry-lifting}
Intuitively, the contact-aware intention embedding $\vf_c$ from the MLLM can capture interaction intention and contact information. However, owing to the inherent limitations of 2D representations, $\vf_c$ lacks the geometric details essential for precise 3D localization. Instead of lifting 2D contact maps to spatial space as in InteractVLM~\cite{dwivedi_interactvlm_2025}, our approach directly improves the 3D awareness of $\vf_c$ by multi-scale geometric features in a general way. This design eliminates the need for camera parameters or intermediate 2D results, offering greater flexibility and applicability.

The 3D features $\{\vf_p^{(i)}\}_{i=1}^R$ from $\gE_{\psi}^{dec}$ provide different levels of geometry, spanning from coarse structure to fine-grained details. We progressively lift the embedding $\vf_c$ by incorporating these geometric cues to enhance its 3D awareness, as demonstrated in \figref{fig:overview} (bottom left). Specifically, for each scale $i$, we form the query, key, and value as:
\begin{equation}
    \vq^{(i)} = \vf_c^{(i-1)} \vw_q, \quad \vk^{(i)} = \vf_p^{(i)} \vw_k, \quad \vv^{(i)} = \vf_p^{(i)} \vw_v
\end{equation}
by projecting $\vf_c^{(i-1)}$ and $\vf_p^{(i)}$ with weights $\vw_q, \vw_k, \vw_v \in \R^{d \times d}$, where $\vf_c^{(0)} = \vf_c$ and $i\in\{1,2,\ldots,R \}$. For simplicity, the index $i$ is omitted for corresponding weights. The correlation between the embedding and geometric features is modeled via an attention mechanism, and the embedding is updated with a residual connection~\cite{he2016deep}:
\begin{equation}
    \vf_c^{'(i)} = \vf_c^{(i-1)} + \texttt{Softmax}\left(\frac{\vq^{(i)} (\vk^{(i)})^T}{\sqrt{d}}\right) \vv^{(i)}.
\end{equation}
The lifted $\vf_c^{(i)}$ at scale $i$ is then obtained by passing through a feed-forward network (FFN) with another residual connection:
\begin{equation}
    \vf_c^{(i)} = \vf_c^{'(i)} + \texttt{FFN}(\vf_c^{'(i)}).
\end{equation}
By progressively integrating multi-scale geometric features, the embedding captures both global shape and local surface characteristics. The final embedding $\vf_c^{3D} = \vf_c^{(R)}$ thus becomes geometrically enriched and fully 3D-aware.

\subsection{Affordance Decoding and Loss Functions}
\label{subsec:affordance-decoding}
We decode the enhanced point-wise features $\tilde{\vf}_p$ into the affordance feature $\vf$ using a point-to-intention attention layer~\cite{yu2024seqafford}, which attends to the 3D-aware intention embedding $\vf_c^{3D}$. The final affordance map $\vp$ is obtained by projecting $\vf$ through an MLP head $\phi_d$ followed by a sigmoid activation $\sigma(\cdot)$:
\begin{equation}
    \vp = \sigma(\phi_d(\vf)).
\end{equation}

During training, we adopt a combined loss for affordance supervision, following the practice in GREAT~\cite{shao2024great}, which includes a focal loss~\cite{lin2017focal} and a dice loss~\cite{milletari2016v}:
\begin{equation}
    \gL_{aff} = \gL_{focal} + \gL_{dice}.
\end{equation}
The overall training objective $\gL$ is a weighted sum of language modeling loss $\gL_{txt}$ and the affordance loss $\gL_{aff}$:
\begin{equation}
    \gL = \lambda_{txt} \gL_{txt} + \lambda_{aff} \gL_{aff},
\end{equation}
where $\lambda_{txt}$ and $\lambda_{aff}$ are balancing coefficients.

\section{Experiment}
\label{sec:exp}

\begin{table*}[!htbp]
    \centering
    \caption{\textbf{Performance comparison on PIAD~\cite{yang2023grounding}.} Best results are highlight in \textbf{bold}. aIOU and AUC are reported in percentage (\%).}
    \tablestyle{1.8pt}{0.9}

\resizebox{0.99\textwidth}{!}{
    \begin{tabular}{x{25mm}|x{25mm}|x{10mm}x{10mm}x{10mm}x{10mm}|x{10mm}x{10mm}x{10mm}x{10mm}}
        \toprule
        \multirow{2}{*}{\textbf{Methods}} & \multirow{2}{*}{\textbf{Venue}} & \multicolumn{4}{c|}{\textbf{Seen}} & \multicolumn{4}{c}{\textbf{Unseen}} \\
        \cmidrule(lr){3-6} \cmidrule(lr){7-10}
        {} & {} & aIOU$\uparrow$ & AUC$\uparrow$ & SIM$\uparrow$ & MAE$\downarrow$ & aIOU$\uparrow$ & AUC$\uparrow$ & SIM$\uparrow$ & MAE$\downarrow$ \\
        \midrule
        \multicolumn{10}{c}{\textit{Intention-Driven 3D Affordance Grounding}} \\
        \midrule
        PMF~\cite{zhuang2021perception} & ICCV 21 & 10.13 & 75.05 & 0.425 & 0.141 & 4.67 & 60.25 & 0.330 & 0.211 \\
        ILN~\cite{chen2022imlovenet} & SIGGRAPH 22 & 11.25 & 75.84 & 0.427 & 0.137 & 4.71 & 59.69 & 0.325 & 0.207 \\
        FRCNN~\cite{xu2023fusionrcnn} & Remote Sensing 23 & 11.97 & 76.05 & 0.429 & 0.136 & 4.71 & 61.92 & 0.332 & 0.195 \\
        PFusion~\cite{xu2018pointfusion} & CVPR 18 & 12.31 & 77.50 & 0.432 & 0.135 & 5.33 & 61.87 & 0.330 & 0.193 \\
        XMFNet~\cite{aiello2022cross} & NeurIPS 22 & 12.94 & 78.25 & 0.441 & 0.127 & 5.68 & 62.58 & 0.342 & 0.188 \\
        IAGNet~\cite{yang2023grounding} & ICCV 23 & \underline{20.51} & 84.85 & 0.545 & 0.098 & 7.95 & \underline{71.84} & \underline{0.352} & 0.127 \\
        GREAT~\cite{shao2024great} & CVPR 25 & 19.61 & \underline{85.22} & \underline{0.569} & \underline{0.093} & \underline{8.32} & 67.46 & 0.330 & \underline{0.121} \\
        \midrule
        \rowcolor{ourscolor} \textbf{\ours (Ours)} & - & \textbf{22.20} & \textbf{88.43} & \textbf{0.605} & \textbf{0.083} & \textbf{13.71} & \textbf{80.92} & \textbf{0.449} & \textbf{0.109} \\
        \textcolor{offsetcolor}{$\Delta$} & - & \textcolor{offsetcolor}{\textbf{+1.69}} & \textcolor{offsetcolor}{\textbf{+3.21}} & \textcolor{offsetcolor}{\textbf{+0.036}} & \textcolor{offsetcolor}{\textbf{+0.010}} & \textcolor{offsetcolor}{\textbf{+5.39}} & \textcolor{offsetcolor}{\textbf{+9.06}} & \textcolor{offsetcolor}{\textbf{+0.097}} & \textcolor{offsetcolor}{\textbf{+0.012}} \\
        \midrule
        \multicolumn{10}{c}{\textcolor{darkgray}{\textit{Language-Driven 3D Affordance Grounding}}} \\
        \midrule
        \textcolor{darkgray}{LASO~\cite{li2024laso}} & \textcolor{darkgray}{CVPR 24} & \textcolor{darkgray}{19.70} & \textcolor{darkgray}{84.20} & \textcolor{darkgray}{0.590} & \textcolor{darkgray}{0.096} & \textcolor{darkgray}{8.00} & \textcolor{darkgray}{69.20} & \textcolor{darkgray}{0.386} & \textcolor{darkgray}{0.118} \\
        \textcolor{darkgray}{GEAL~\cite{lu2024geal}} & \textcolor{darkgray}{CVPR 25} & \textcolor{darkgray}{22.50} & \textcolor{darkgray}{85.00} & \textcolor{darkgray}{0.600} & \textcolor{darkgray}{0.092} & \textcolor{darkgray}{8.70} & \textcolor{darkgray}{72.50} & \textcolor{darkgray}{0.390} & \textcolor{darkgray}{0.102} \\
        \bottomrule
    \end{tabular}
}
    \label{tab:piadv1}
\end{table*}

\begin{table*}[!htbp]
    \centering
    \caption{\textbf{Performance comparison on PIADv2~\cite{shao2024great}.} Best results are highlight in \textbf{bold}. aIOU and AUC are reported in percentage (\%).}
    \tablestyle{1.5pt}{0.9}
    \resizebox{0.99\textwidth}{!}{
    \begin{tabular}{x{20mm}|x{10mm}|x{8mm}x{8mm}x{8mm}x{8mm}|x{8mm}x{8mm}x{8mm}x{8mm}|x{8mm}x{8mm}x{8mm}x{8mm}}
        \toprule
        \multirow{2}{*}{\textbf{Methods}} & \multirow{2}{*}{\textbf{Year}} & \multicolumn{4}{c|}{\textbf{Seen}} & \multicolumn{4}{c|}{\textbf{Unseen Object}} & \multicolumn{4}{c}{\textbf{Unseen Affordance}} \\
        \cmidrule(lr){3-6} \cmidrule(lr){7-10} \cmidrule(lr){11-14}
        {} & {} & aIOU$\uparrow$ & AUC$\uparrow$ & SIM$\uparrow$ & MAE$\downarrow$ & aIOU$\uparrow$ & AUC$\uparrow$ & SIM$\uparrow$ & MAE$\downarrow$ & aIOU$\uparrow$ & AUC$\uparrow$ & SIM$\uparrow$ & MAE$\downarrow$ \\
        \midrule
        OpenAD~\cite{nguyen2023open} & 2023 & 31.88 & 89.54 & 0.526 & 0.104 & 16.62 & 73.49 & 0.339 & 0.159 & 8.00 & 61.22 & 0.229 & 0.167 \\
        FRCNN~\cite{xu2023fusionrcnn} & 2023 & 33.55 & 87.05 & 0.600 & 0.082 & 18.08 & 72.20 & 0.362 & 0.152 & 7.96 & 59.08 & 0.210 & 0.156 \\
        XMFNet~\cite{aiello2022cross} & 2022 & 33.91 & 87.39 & 0.604 & 0.078 & 17.40 & 74.61 & 0.361 & 0.126 & 8.11 & 60.99 & 0.225 & 0.152 \\
        IAGNet~\cite{yang2023grounding} & 2024 & 34.29 & 89.03 & 0.623 & 0.076 & 16.78 & 73.03 & 0.351 & 0.123 & 8.99 & 62.29 & 0.251 & 0.141 \\
        LASO~\cite{li2024laso} & 2024 & 34.88 & 90.34 & 0.627 & 0.077 & 16.05 & 73.32 & 0.354 & 0.123 & 8.37 & 64.07 & 0.228 & 0.140 \\
        GREAT~\cite{shao2024great} & 2025 & \underline{37.61} & \underline{91.24} & \underline{0.660} & \underline{0.067} & \underline{19.16} & \underline{76.90} & \underline{0.384} & \underline{0.119} & \underline{12.78} & \underline{69.26} & \underline{0.297} & \underline{0.135} \\
        \midrule
        \rowcolor{ourscolor} \textbf{\ours} & - & \textbf{40.06} & \textbf{94.19} & \textbf{0.698} & \textbf{0.063} & \textbf{24.28} & \textbf{84.78} & \textbf{0.449} & \textbf{0.112} & \textbf{13.28} & \textbf{72.21} & \textbf{0.302} & \textbf{0.128} \\
        \bottomrule
    \end{tabular}
}
    \label{tab:paidv2}
\end{table*}

\begin{figure*}[!htbp]
    \centering
    \includegraphics[width=0.99\linewidth]{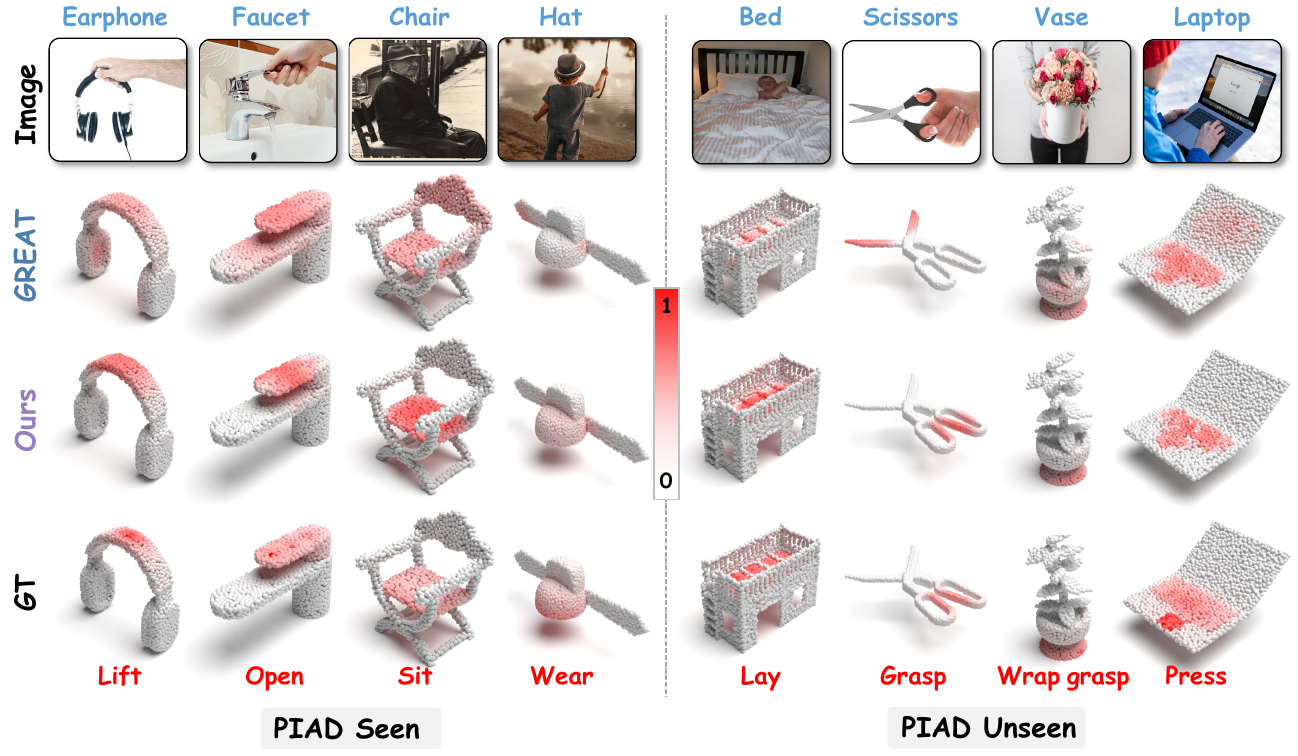}
    \vspace{-0.5em}
    \caption{\textbf{Qualitative comparison on PIAD~\cite{yang2023grounding}.} Our \ours~generates more precise and complete affordance predictions compared to GREAT~\cite{shao2024great}, highlighting its enhanced capability in understanding interaction intentions and reasoning about 3D affordances.}
    \label{fig:piadv1}
\end{figure*}

\subsection{Experimental Settings}

\mypara{Datasets and Metrics.} We evaluate our method on two standard point-image affordance datasets: PIAD~\cite{yang2023grounding} and PIADv2~\cite{shao2024great}. \textit{\textbf{PIAD}} includes over 7k point clouds spanning 23 object categories and 17 affordance types, while \textit{\textbf{PIADv2}} extends this to 38.8k object instances with 43 categories and 24 affordance types. PIADv2 is organized into three partitions: \textit{Seen}, \textit{Unseen Object}, and \textit{Unseen Affordance}. The Seen split shares the same objects and affordances between training and testing, while the Unseen Object and Unseen Affordance splits contain novel object types and novel affordance categories not present during training, respectively. PIAD merges the latter two splits into a single \textit{Unseen} partition. To further evaluate the robustness, we construct a corrupted benchmark by introducing various noise types to point clouds, building upon GEAL~\cite{lu2024geal} and PIAD~\cite{yang2023grounding}. Following established evaluation protocols~\cite{yang2023grounding,li2024laso,shao2024great,dwivedi_interactvlm_2025}, we report four metrics: average Interaction Overlap (\textbf{aIOU}), Area Under the ROC Curve (\textbf{AUC}), Similarity (\textbf{SIM}), and Mean Absolute Error (\textbf{MAE}).

\mypara{Implementation Details.} Our \ours framework integrates the 2D MLLM Qwen2.5-VL~\cite{bai2025qwen2} with the 3D backbone PointNet++~\cite{qi2017pointnet++}. The model is trained end-to-end using mixed BF16 precision, with the point backbone maintained in full precision for stable optimization. We apply LoRA~\cite{hu2022lora} fine-tuning to the language component of the MLLM using a rank of 16. Training is performed on 4 NVIDIA H20 GPUs with a global batch size of 64. We use AdamW~\cite{loshchilov2017decoupled} as the optimizer, with an initial learning rate of $1e-4$ and a linear learning rate scheduler. The loss weights are set to $\lambda_{txt} = 1.0$ and $\lambda_{aff} = 2.0$.

\subsection{Main Results}
\mypara{Evaluation on PIAD and PIADv2.}
\tabref{tab:piadv1} and~\tabref{tab:paidv2} compare the performance of \ours against current state-of-the-art works, including intention-driven~\cite{shao2024great,dwivedi_interactvlm_2025,yu2024seqafford} and language-driven~\cite{li2024laso,lu2024geal} approaches. As shown in~\tabref{tab:piadv1}, our method consistently outperforms other intention-driven counterparts across both seen and unseen splits. Specifically, \ours surpasses GREAT~\cite{shao2024great} and IAGNet~\cite{yang2023grounding} by 2.59\% and 1.69\% in aIOU on the seen subset, respectively. As for the unseen split, \ours achieves a substantial improvement of 5.39\% in aIOU over GREAT, demonstrating superior generalization to novel objects and affordances. Our approach also exhibits advantages over language-driven methods such as LASO~\cite{li2024laso} and GEAL~\cite{lu2024geal} on most metrics, particularly in AUC and SIM. On the more extensive PIADv2 benchmark (\tabref{tab:paidv2}), \ours obtains the best performance across all three splits, exceeding GREAT by 2.45\%, 5.12\%, and 0.5\% in aIOU for Seen, Unseen Object, and Unseen Affordance, respectively. The significant gain on unseen objects highlights our architecture's capability to comprehensively understand novel object geometries. Although the improvement for unseen affordances is more modest, this is expected given the inherent difficulty of generalizing to entirely new interaction types without prior exposure. These results collectively validate the effectiveness of our framework in handling diverse and complex affordance grounding scenarios.

\begin{figure}
    \centering
    \includegraphics[width=0.99\linewidth]{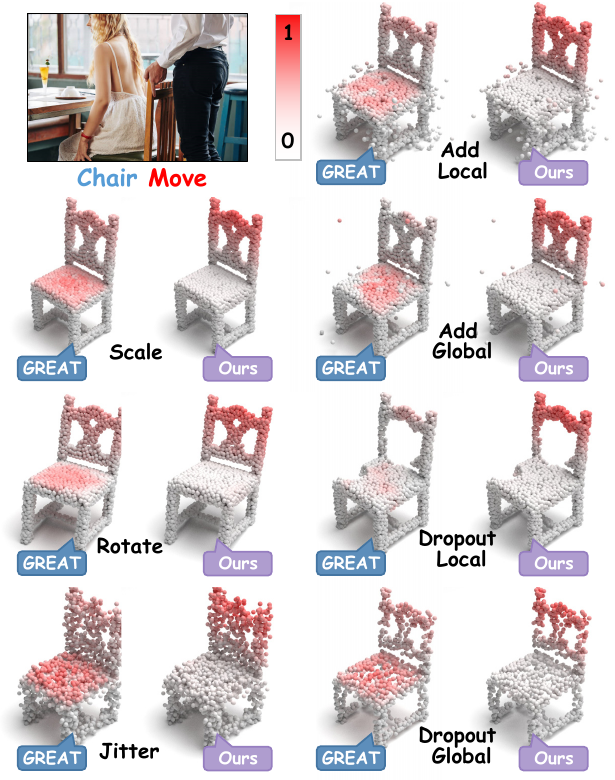}
    \vspace{-0.3em}
    \caption{\textbf{Qualitative comparison on corrupted point clouds.} The proposed \ours~realizes stronger robustness in accurately localizing affordance regions under severe noise corruption, while GREAT~\cite{shao2024great} struggles to maintain reliable predictions.}
    \label{fig:robustness}
    \vspace{-1em}
\end{figure}

\mypara{Robustness Assessment.}
We further validate the robustness of \ours on a corrupted benchmark constructed following GEAL~\cite{lu2024geal} by introducing various noise types to point clouds. Please see~\secref{sec:corrupted-benchmark} in the supplementary material for detailed benchmark construction. As shown in~\tabref{tab:geal}, our method achieves superior performance than GREAT~\cite{shao2024great} under various corruption styles, indicating robust affordance grounding in challenging conditions. Notably, \ours outperforms GREAT by significant margins of 5.69\%, 9.31\%, and 6.17\% in aIOU under jitter, dropout-local, and add-local corruptions, respectively.

\begin{table}[t]
    \centering
    \caption{\textbf{Robustness evaluation.} \textit{Drop}, \textit{L} and \textit{G} indicate dropout, local and global corruptions, respectively.}
    \tablestyle{1.4pt}{1.1}
    \resizebox{0.48\textwidth}{!}{
    \begin{tabular}{y{10mm}|x{9mm}x{7mm}|x{9mm}x{7mm}|x{9mm}x{7mm}|x{9mm}x{7mm}}
        \toprule
        \multirow{2}{*}{\textbf{Type}} & \multicolumn{2}{c|}{\textbf{aIOU}$\uparrow$} & \multicolumn{2}{c|}{\textbf{AUC}$\uparrow$} & \multicolumn{2}{c|}{\textbf{SIM}$\uparrow$} & \multicolumn{2}{c}{\textbf{MAE}$\downarrow$} \\
         & GREAT & \cellcolor{ourscolor} \textbf{Ours} & GREAT & \cellcolor{ourscolor} \textbf{Ours} & GREAT & \cellcolor{ourscolor} \textbf{Ours} & GREAT & \cellcolor{ourscolor} \textbf{Ours} \\
        \midrule
        \textbf{Scale} & 13.41 & \cellcolor{ourscolor} \textbf{19.18} & 77.77 & \cellcolor{ourscolor} \textbf{85.73} & 0.490 & \cellcolor{ourscolor} \textbf{0.596} & 0.116 & \cellcolor{ourscolor} \textbf{0.093} \\
        \textbf{Jitter} & 11.67 & \cellcolor{ourscolor} \textbf{17.36} & 75.65 & \cellcolor{ourscolor} \textbf{84.57} & 0.461 & \cellcolor{ourscolor} \textbf{0.568} & 0.115 & \cellcolor{ourscolor} \textbf{0.095} \\
        \textbf{Rotate} & 12.61 & \cellcolor{ourscolor} \textbf{18.33} & 76.99 & \cellcolor{ourscolor} \textbf{85.31} & 0.483 & \cellcolor{ourscolor} \textbf{0.591} & 0.116 & \cellcolor{ourscolor} \textbf{0.095} \\
        \textbf{Drop-L} & 8.67 & \cellcolor{ourscolor} \textbf{17.98} & 72.38 & \cellcolor{ourscolor} \textbf{84.59} & 0.420 & \cellcolor{ourscolor} \textbf{0.568} & 0.122 & \cellcolor{ourscolor} \textbf{0.098} \\
        \textbf{Drop-G} & 13.62 & \cellcolor{ourscolor} \textbf{20.38} & 78.26 & \cellcolor{ourscolor} \textbf{86.76} & 0.494 & \cellcolor{ourscolor} \textbf{0.611} & 0.112 & \cellcolor{ourscolor} \textbf{0.089} \\
        \textbf{Add-L} & 11.50 & \cellcolor{ourscolor} \textbf{17.67} & 75.16 & \cellcolor{ourscolor} \textbf{86.04} & 0.438 & \cellcolor{ourscolor} \textbf{0.552} & 0.109 & \cellcolor{ourscolor} \textbf{0.093} \\
        \textbf{Add-G} & 9.63 & \cellcolor{ourscolor} \textbf{19.09} & 71.25 & \cellcolor{ourscolor} \textbf{86.36} & 0.432 & \cellcolor{ourscolor} \textbf{0.578} & 0.108 & \cellcolor{ourscolor} \textbf{0.090} \\
        \bottomrule
    \end{tabular}
}

    \label{tab:geal}
    \vspace{-1em}
\end{table}

\subsection{Ablation Studies and Analysis}
This section presents extensive ablation studies conducted on PIAD to verify the efficacy of our proposed components and provide an in-depth analysis of our design choices.

\begin{table}[!htbp]
    \centering
    \caption{\textbf{Ablation study on affordance-guided intention embedding.} \textit{w/ cls} and \textit{w/ afford} denote including object category in the prompt and predicting affordance label in the output, respectively.}
    \tablestyle{2pt}{1.0}
    
\resizebox{0.49\textwidth}{!}{
    \begin{tabular}
    {x{8mm}|x{9mm}|x{12mm}|x{8mm}x{8mm}x{8mm}x{8mm}}
        \toprule
        \textbf{Type} & \textit{w/ \textbf{cls}} & \textit{w/ \textbf{afford}} & \textbf{aIOU}$\uparrow$ & \textbf{AUC}$\uparrow$ & \textbf{SIM}$\uparrow$ & \textbf{MAE}$\downarrow$ \\
        \midrule
        \multirow{3}{*}{\rotatebox{90}{Seen}} & \cellcolor{ourscolor} \YesV & \cellcolor{ourscolor} \YesV & \cellcolor{ourscolor} \textbf{22.20} & \cellcolor{ourscolor} \textbf{88.43} & \cellcolor{ourscolor} \textbf{0.605} & \cellcolor{ourscolor} \textbf{0.083} \\
         & \NoX & \YesV & 21.95 & 87.73 & 0.594 & 0.085 \\
         & \NoX & \NoX & 19.72 & 84.36 & 0.564 & 0.094 \\
        \midrule
        \multirow{3}{*}{\rotatebox{90}{Unseen}} & \cellcolor{ourscolor} \YesV & \cellcolor{ourscolor} \YesV & \cellcolor{ourscolor} \textbf{13.71} & \cellcolor{ourscolor} \textbf{80.92} & \cellcolor{ourscolor} \textbf{0.449} & \cellcolor{ourscolor} \textbf{0.109} \\
         & \NoX & \YesV & 12.35 & 79.28 & 0.425 & 0.113 \\
         & \NoX & \NoX & 8.78 & 66.33 & 0.345 & 0.132 \\
        \bottomrule
    \end{tabular}
}
    \label{tab:prompt_ablation}
    \vspace{-1em}
\end{table}

\begin{table}[!htbp]
    \centering
    \caption{\textbf{Ablation study of core components.} \textit{Hier. Integ.} and \textit{Geo. Lifting} mean the hierarchical cross-modal integration and multi-granular geometry lifting, respectively.}
    \tablestyle{2.4pt}{1.0}
    \resizebox{0.49\textwidth}{!}{
    \begin{tabular}{x{6mm}|y{19mm}|x{10mm}x{10mm}x{10mm}x{10mm}}
        \toprule
        \textbf{Type} & \textbf{Ablation} & \textbf{aIOU}$\uparrow$ & \textbf{AUC}$\uparrow$ & \textbf{SIM}$\uparrow$ & \textbf{MAE}$\downarrow$ \\
        \midrule
        \multirow{5}{*}{\rotatebox{90}{Seen}} & \cellcolor{ourscolor} \textbf{\ours} & \cellcolor{ourscolor} \textbf{22.20} & \cellcolor{ourscolor} \textbf{88.43} & \cellcolor{ourscolor} \textbf{0.605} & \cellcolor{ourscolor} \textbf{0.083} \\
         & \textit{w/o Hier. Integ.} & 21.15 & 87.64 & 0.597 & 0.085 \\
         & \textit{w/o Geo. Lifting} & 20.46 & 87.25 & 0.588 & 0.087 \\
         & \textit{w/o Both} & 19.55 & 86.67 & 0.587 & 0.088 \\
         & \textit{w/o $\gL_{txt}$} & 20.69 & 87.66 & 0.603 & 0.086 \\
        \midrule
        \multirow{5}{*}{\rotatebox{90}{Unseen}} & \cellcolor{ourscolor} \textbf{\ours} & \cellcolor{ourscolor} \textbf{13.71} & \cellcolor{ourscolor} \textbf{80.92} & \cellcolor{ourscolor} \textbf{0.449} & \cellcolor{ourscolor} \textbf{0.109} \\
         & \textit{w/o Hier. Integ.} & 10.50 & 74.72 & 0.407 & 0.116 \\
         & \textit{w/o Geo. Lifting} & 10.20 & 77.71 & 0.418 & 0.115 \\
         & \textit{w/o Both} & 7.86 & 75.88 & 0.393 & 0.123 \\
         & \textit{w/o $\gL_{txt}$} & 9.78 & 74.27 & 0.412 & 0.111 \\
        \bottomrule
    \end{tabular}
}
    \label{tab:ablation_components}
    \vspace{-1.5em}
\end{table}

\begin{figure}
    \centering
    \includegraphics[width=0.99\linewidth]{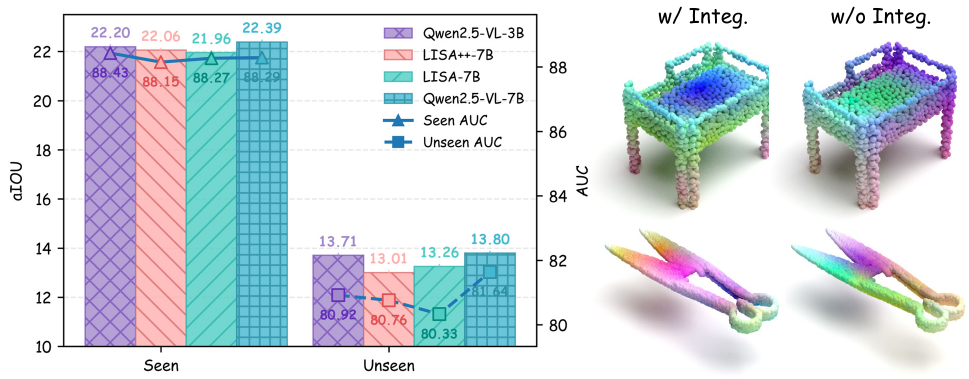}
    \vspace{-0.3em}
    \caption{Impact of MLLM backbones (\textit{left}) and visualization of point-wise features via PCA (\textit{right}).}
    \label{fig:mllm}
\end{figure}

\mypara{Core Components Ablation.}
We first examine the impact of our affordance-guided intention embedding strategy in~\tabref{tab:prompt_ablation}. We gradually ablate the object-centric prompt strategy by removing the category from the prompt and disabling textual affordance label prediction in the MLLM output. Results indicate that excluding category information (\textit{w/o cls}) and affordance labels (\textit{w/o afford}) leads to performance degradation of 2.49\% and 4.93\% in aIOU for the seen and unseen splits, respectively, demonstrating their importance in capturing object and contextual cues. We further evaluate the hierarchical cross-modal integration and multi-granular geometry lifting modules in~\tabref{tab:ablation_components}. Removing the integration mechanism (\textit{w/o Hier. Integ.}) results in decreased performance across all metrics, validating its effectiveness in fusing complementary information for improved representation alignment. To visualize this effect, we apply PCA to reduce the dimension of integrated features (see~\figref{fig:mllm} \textit{right} and supplementary figure~\figref{fig:sup_integ}). The enhanced features exhibit clearer clustering patterns, indicating improved discriminative capability for different regions. Eliminating the geometry lifting module (\textit{w/o Geo. Lifting}) also causes significant performance degradation, with aIOU dropping by 1.74\% and 3.51\% on the seen and unseen splits, respectively, underscoring its crucial role in enriching geometric awareness. The absence of both components (\textit{w/o Both}) leads to the most pronounced performance decline, confirming their complementary contributions in achieving accurate 3D affordance grounding. Additionally, we find the textual loss $\gL_{txt}$ also contributes to performance.

\mypara{Sensitivity to MLLMs.} We investigate the impact of different MLLM backbones in ~\figref{fig:mllm} \textit{left}. Although LISA~\cite{lai2024lisa} and LISA++~\cite{yang2023lisa++} are specifically fine-tuned for segmentation tasks, Qwen2.5-VL~\cite{bai2025qwen2} still achieves superior performance, particularly on the unseen split. This suggests that robust visual-language understanding and broader world knowledge are more beneficial for 3D affordance grounding than specialized segmentation capabilities. While the Qwen2.5-VL-7B yields slightly higher aIOU than the 3B version, we select the latter to strike a balance between performance and computational efficiency. More experiments are provided in~\secref{sec:additional-exp} of the supplementary material.

\subsection{Qualitative Results}
To illustrate the effectiveness and robustness of \ours, we present visualization comparisons on PIAD~\cite{yang2023grounding} and our constructed corrupted benchmark. \figref{fig:piadv1} shows several representative examples comparing our method with GREAT~\cite{shao2024great} against ground truth annotations across both seen and unseen object splits. The visual results indicate that \ours produces more precise and complete affordance predictions compared to GREAT. For instance, our method accurately identifies the operable region of a faucet handle, while GREAT incorrectly attributes affordance to larger portions of the faucet body. In the case of unseen objects, \ours correctly localizes the press area of a laptop, but GREAT yields fragmented and inaccurate predictions. Furthermore, \figref{fig:robustness} presents results on point clouds corrupted with various noise patterns. Under challenging conditions such as coordinate jitter and local point dropout, \ours maintains stable and accurate affordance localization, while GREAT exhibits performance degradation with inconsistent and unreliable outputs. These qualitative comparisons substantiate the superior effectiveness and resilience of our framework in real-world scenarios with imperfect data. Please refer to~\secref{sec:more-vis} in the supplementary material for more visualization results.
\section{Conclusion}
\label{sec:conclusion}
In this paper, we present \ours, an innovative framework empowered by an MLLM for intention-driven 3D affordance grounding. Our affordance-guided intention embedding technique effectively captures interaction-related information and contextual cues from the input image, providing essential priors for affordance prediction. The proposed hierarchical cross-modal integration mechanism incorporates sufficient cross-modal knowledge into point cloud features for enhanced object representation alignment, while the multi-granular geometry lifting enriches 3D awareness of the intention embedding with detailed geometric cues for precise spatial localization. Extensive experiments on standard benchmarks and our newly constructed corrupted dataset demonstrate the superiority and robustness of our method. Comprehensive ablation studies systematically validate the effectiveness of each component. For future work, we plan to extend our framework to scene-level affordance grounding tasks, aiming to handle more complex environments and diverse interactions.

\section*{Acknowledgement}
The research work described in this paper was conducted in the JC STEM Lab of Machine Learning and Computer Vision funded by The Hong Kong Jockey Club Charities Trust. This research received partially support from the Global STEM Professorship Scheme from the Hong Kong Special Administrative Region.
{
    \small
    \bibliographystyle{ieeenat_fullname}
    \bibliography{main}

@inproceedings{shao2024great,
  title={GREAT: Geometry-Intention Collaborative Inference for Open-Vocabulary 3D Object Affordance Grounding},
  author={Shao, Yawen and Zhai, Wei and Yang, Yuhang and Luo, Hongchen and Cao, Yang and Zha, Zheng-Jun},
  booktitle={Proceedings of the IEEE/CVF Conference on Computer Vision and Pattern Recognition},
  year={2025}
}

@inproceedings{kim2025david,
  title={DAViD: Modeling Dynamic Affordance of 3D Objects using Pre-trained Video Diffusion Models},
  author={Kim, Hyeonwoo and Beak, Sangwon and Joo, Hanbyul},
  booktitle={Proceedings of the IEEE/CVF International Conference on Computer Vision},
  year={2025}
}

@article{yang2025egochoir,
  title={Egochoir: Capturing 3d human-object interaction regions from egocentric views},
  author={Yang, Yuhang and Zhai, Wei and Wang, Chengfeng and Yu, Chengjun and Cao, Yang and Zha, Zheng-Jun},
  journal={Advances in Neural Information Processing Systems},
  volume={37},
  pages={54529--54557},
  year={2025}
}

@inproceedings{yang2023grounding,
  title={Grounding 3d object affordance from 2d interactions in images},
  author={Yang, Yuhang and Zhai, Wei and Luo, Hongchen and Cao, Yang and Luo, Jiebo and Zha, Zheng-Jun},
  booktitle={Proceedings of the IEEE/CVF International Conference on Computer Vision},
  pages={10905--10915},
  year={2023}
}

@inproceedings{gao2024learning,
  title={Learning 2d invariant affordance knowledge for 3d affordance grounding},
  author={Gao, Xianqiang and Zhang, Pingrui and Qu, Delin and Wang, Dong and Wang, Zhigang and Ding, Yan and Zhao, Bin},
  booktitle={Proceedings of the AAAI Conference on Artificial Intelligence},
  year={2025}
}

@inproceedings{chu20253d,
  title={3D-AffordanceLLM: Harnessing Large Language Models for Open-Vocabulary Affordance Detection in 3D Worlds},
  author={Chu, Hengshuo and Deng, Xiang and Chen, Xiaoyang and Li, Yinchuan and Hao, Jianye and Nie, Liqiang},
  booktitle={International Conference on Learning Representations},
  year={2025}
}

@inproceedings{lu2024geal,
  title={GEAL: Generalizable 3D Affordance Learning with Cross-Modal Consistency},
  author={Lu, Dongyue and Kong, Lingdong and Huang, Tianxin and Lee, Gim Hee},
  booktitle={Proceedings of the IEEE/CVF Conference on Computer Vision and Pattern Recognition},
  year={2025}
}

@inproceedings{yu2024seqafford,
  title={SeqAfford: Sequential 3D Affordance Reasoning via Multimodal Large Language Model},
  author={Yu, Chunlin and Wang, Hanqing and Shi, Ye and Luo, Haoyang and Yang, Sibei and Yu, Jingyi and Wang, Jingya},
  booktitle={Proceedings of the IEEE/CVF Conference on Computer Vision and Pattern Recognition},
  year={2025}
}

@inproceedings{li2024laso,
  title={Laso: Language-guided affordance segmentation on 3d object},
  author={Li, Yicong and Zhao, Na and Xiao, Junbin and Feng, Chun and Wang, Xiang and Chua, Tat-seng},
  booktitle={Proceedings of the IEEE/CVF Conference on Computer Vision and Pattern Recognition},
  pages={14251--14260},
  year={2024}
}

@inproceedings{yang2024lemon,
  title={Lemon: Learning 3d human-object interaction relation from 2d images},
  author={Yang, Yuhang and Zhai, Wei and Luo, Hongchen and Cao, Yang and Zha, Zheng-Jun},
  booktitle={Proceedings of the IEEE/CVF Conference on Computer Vision and Pattern Recognition},
  pages={16284--16295},
  year={2024}
}

@inproceedings{kim2024beyond,
  title={Beyond the contact: Discovering comprehensive affordance for 3d objects from pre-trained 2d diffusion models},
  author={Kim, Hyeonwoo and Han, Sookwan and Kwon, Patrick and Joo, Hanbyul},
  booktitle={European Conference on Computer Vision},
  pages={400--419},
  year={2024},
  organization={Springer}
}

@article{liu2024pavlm,
  title={PAVLM: Advancing Point Cloud based Affordance Understanding Via Vision-Language Model},
  author={Liu, Shang-Ching and Tran, Van Nhiem and Chen, Wenkai and Cheng, Wei-Lun and Huang, Yen-Lin and Liao, I-Bin and Li, Yung-Hui and Zhang, Jianwei},
  journal={arXiv preprint arXiv:2410.11564},
  year={2024}
}

@inproceedings{delitzas2024scenefun3d,
  title={SceneFun3D: fine-grained functionality and affordance understanding in 3D scenes},
  author={Delitzas, Alexandros and Takmaz, Ayca and Tombari, Federico and Sumner, Robert and Pollefeys, Marc and Engelmann, Francis},
  booktitle={Proceedings of the IEEE/CVF Conference on Computer Vision and Pattern Recognition},
  pages={14531--14542},
  year={2024}
}

@inproceedings{van2024open,
  title={Open-vocabulary affordance detection using knowledge distillation and text-point correlation},
  author={Van Vo, Tuan and Vu, Minh Nhat and Huang, Baoru and Nguyen, Toan and Le, Ngan and Vo, Thieu and Nguyen, Anh},
  booktitle={IEEE International Conference on Robotics and Automation (ICRA)},
  pages={13968--13975},
  year={2024},
  organization={IEEE}
}

@inproceedings{nguyen2023open,
  title={Open-vocabulary affordance detection in 3d point clouds},
  author={Nguyen, Toan and Vu, Minh Nhat and Vuong, An and Nguyen, Dzung and Vo, Thieu and Le, Ngan and Nguyen, Anh},
  booktitle={IEEE/RSJ International Conference on Intelligent Robots and Systems (IROS)},
  pages={5692--5698},
  year={2023},
  organization={IEEE}
}

@inproceedings{dwivedi_interactvlm_2025,
  title={InteractVLM: 3D Interaction Reasoning from 2D Foundational Models},
  author={Dwivedi, Sai Kumar and Antić, Dimitrije and Tripathi, Shashank and Taheri, Omid and Schmid, Cordelia and Black, Michael J. and Tzionas, Dimitrios},
  booktitle={Proceedings of the IEEE/CVF Conference on Computer Vision and Pattern Recognition},
  year={2025},
}

@inproceedings{tang2025uad,
  title={UAD: Unsupervised Affordance Distillation for Generalization in Robotic Manipulation},
  author={Tang, Yihe and Huang, Wenlong and Wang, Yingke and Li, Chengshu and Yuan, Roy and Zhang, Ruohan and Wu, Jiajun and Fei-Fei, Li},
  booktitle={IEEE International Conference on Robotics and Automation (ICRA)},
  year={2025}
}

@article{ma2024glover,
  title={Glover: Generalizable open-vocabulary affordance reasoning for task-oriented grasping},
  author={Ma, Teli and Wang, Zifan and Zhou, Jiaming and Wang, Mengmeng and Liang, Junwei},
  journal={arXiv preprint arXiv:2411.12286},
  year={2024}
}

@article{xu2023fusionrcnn,
  title={Fusionrcnn: Lidar-camera fusion for two-stage 3d object detection},
  author={Xu, Xinli and Dong, Shaocong and Xu, Tingfa and Ding, Lihe and Wang, Jie and Jiang, Peng and Song, Liqiang and Li, Jianan},
  journal={Remote Sensing},
  volume={15},
  number={7},
  pages={1839},
  year={2023},
  publisher={MDPI}
}

@article{aiello2022cross,
  title={Cross-modal learning for image-guided point cloud shape completion},
  author={Aiello, Emanuele and Valsesia, Diego and Magli, Enrico},
  journal={Advances in Neural Information Processing Systems},
  volume={35},
  pages={37349--37362},
  year={2022}
}

@inproceedings{chen2022imlovenet,
  title={Imlovenet: Misaligned image-supported registration network for low-overlap point cloud pairs},
  author={Chen, Honghua and Wei, Zeyong and Xu, Yabin and Wei, Mingqiang and Wang, Jun},
  booktitle={ACM SIGGRAPH 2022 conference proceedings},
  pages={1--9},
  year={2022}
}

@inproceedings{xu2018pointfusion,
  title={Pointfusion: Deep sensor fusion for 3d bounding box estimation},
  author={Xu, Danfei and Anguelov, Dragomir and Jain, Ashesh},
  booktitle={Proceedings of the IEEE/CVF Conference on Computer Vision and Pattern Recognition},
  pages={244--253},
  year={2018}
}

@inproceedings{zhuang2021perception,
  title={Perception-aware multi-sensor fusion for 3d lidar semantic segmentation},
  author={Zhuang, Zhuangwei and Li, Rong and Jia, Kui and Wang, Qicheng and Li, Yuanqing and Tan, Mingkui},
  booktitle={Proceedings of the IEEE/CVF International Conference on Computer Vision},
  pages={16280--16290},
  year={2021}
}

@article{qi2017pointnet++,
  title={Pointnet++: Deep hierarchical feature learning on point sets in a metric space},
  author={Qi, Charles Ruizhongtai and Yi, Li and Su, Hao and Guibas, Leonidas J},
  journal={Advances in Neural Information Processing Systems},
  volume={30},
  publisher={IEEE},
  year={2017}
}

@article{bai2025qwen2,
  title={Qwen2. 5-vl technical report},
  author={Bai, Shuai and Chen, Keqin and Liu, Xuejing and Wang, Jialin and Ge, Wenbin and Song, Sibo and Dang, Kai and Wang, Peng and Wang, Shijie and Tang, Jun and others},
  journal={arXiv preprint arXiv:2502.13923},
  year={2025}
}

@inproceedings{hu2022lora,
  title={Lora: Low-rank adaptation of large language models.},
  author={Hu, Edward J and Shen, Yelong and Wallis, Phillip and Allen-Zhu, Zeyuan and Li, Yuanzhi and Wang, Shean and Wang, Lu and Chen, Weizhu and others},
  booktitle={International Conference on Learning Representations},
  year={2022}
}

@article{loshchilov2017decoupled,
  title={Decoupled weight decay regularization},
  author={Loshchilov, Ilya and Hutter, Frank},
  journal={arXiv preprint arXiv:1711.05101},
  year={2017}
}

@inproceedings{kirillov2023segment,
  title={Segment anything},
  author={Kirillov, Alexander and Mintun, Eric and Ravi, Nikhila and Mao, Hanzi and Rolland, Chloe and Gustafson, Laura and Xiao, Tete and Whitehead, Spencer and Berg, Alexander C and Lo, Wan-Yen and others},
  booktitle={Proceedings of the IEEE/CVF International Conference on Computer Vision},
  pages={4015--4026},
  year={2023}
}

@inproceedings{radford2021learning,
  title={Learning transferable visual models from natural language supervision},
  author={Radford, Alec and Kim, Jong Wook and Hallacy, Chris and Ramesh, Aditya and Goh, Gabriel and Agarwal, Sandhini and Sastry, Girish and Askell, Amanda and Mishkin, Pamela and Clark, Jack and others},
  booktitle={International Conference on Machine Learning},
  pages={8748--8763},
  year={2021}
}

@article{kerbl20233d,
  title={3d gaussian splatting for real-time radiance field rendering.},
  author={Kerbl, Bernhard and Kopanas, Georgios and Leimk{\"u}hler, Thomas and Drettakis, George},
  volume={42},
  number={4},
  pages={139--1},
  journal={ACM Transactions on Graphics},
  year={2023}
}

@article{yao2025gaussiancross,
  title={GaussianCross: Cross-modal Self-supervised 3D Representation Learning via Gaussian Splatting},
  author={Yao, Lei and Wang, Yi and Zhang, Yi and Liu, Moyun and Chau, Lap-Pui},
  journal={arXiv preprint arXiv:2508.02172},
  year={2025}
}

@article{touvron2023llama,
  title={Llama: Open and efficient foundation language models},
  author={Touvron, Hugo and Lavril, Thibaut and Izacard, Gautier and Martinet, Xavier and Lachaux, Marie-Anne and Lacroix, Timoth{\'e}e and Rozi{\`e}re, Baptiste and Goyal, Naman and Hambro, Eric and Azhar, Faisal and others},
  journal={arXiv preprint arXiv:2302.13971},
  year={2023}
}

@inproceedings{qi2024shapellm,
  title={Shapellm: Universal 3d object understanding for embodied interaction},
  author={Qi, Zekun and Dong, Runpei and Zhang, Shaochen and Geng, Haoran and Han, Chunrui and Ge, Zheng and Yi, Li and Ma, Kaisheng},
  booktitle={European Conference on Computer Vision},
  pages={214--238},
  year={2024},
  organization={Springer}
}

@inproceedings{lai2024lisa,
  title={Lisa: Reasoning segmentation via large language model},
  author={Lai, Xin and Tian, Zhuotao and Chen, Yukang and Li, Yanwei and Yuan, Yuhui and Liu, Shu and Jia, Jiaya},
  booktitle={Proceedings of the IEEE/CVF Conference on Computer Vision and Pattern Recognition},
  pages={9579--9589},
  year={2024}
}

@inproceedings{wu2025afforddp,
  title={Afforddp: Generalizable diffusion policy with transferable affordance},
  author={Wu, Shijie and Zhu, Yihang and Huang, Yunao and Zhu, Kaizhen and Gu, Jiayuan and Yu, Jingyi and Shi, Ye and Wang, Jingya},
  booktitle={Proceedings of the Computer Vision and Pattern Recognition Conference},
  pages={6971--6980},
  year={2025}
}

@inproceedings{bahl2023affordances,
  title={Affordances from human videos as a versatile representation for robotics},
  author={Bahl, Shikhar and Mendonca, Russell and Chen, Lili and Jain, Unnat and Pathak, Deepak},
  booktitle={Proceedings of the IEEE/CVF Conference on Computer Vision and Pattern Recognition},
  pages={13778--13790},
  year={2023}
}

@article{quesada2022proactive,
  title={Proactive robot assistance: Affordance-aware augmented reality user interfaces},
  author={Quesada, Rodrigo Chac{\'o}n and Demiris, Yiannis},
  journal={IEEE Robotics \& Automation Magazine},
  volume={29},
  number={1},
  pages={22--34},
  year={2022},
  publisher={IEEE}
}

@article{vaswani2017attention,
  title={Attention is all you need},
  author={Vaswani, Ashish and Shazeer, Noam and Parmar, Niki and Uszkoreit, Jakob and Jones, Llion and Gomez, Aidan N and Kaiser, {\L}ukasz and Polosukhin, Illia},
  journal={Advances in Neural Information Processing Systems},
  volume={30},
  year={2017}
}

@inproceedings{he2016deep,
  title={Deep residual learning for image recognition},
  author={He, Kaiming and Zhang, Xiangyu and Ren, Shaoqing and Sun, Jian},
  booktitle={Proceedings of the IEEE conference on computer vision and pattern recognition},
  pages={770--778},
  year={2016}
}

@article{yao2024sgiformer,
  title={SGIFormer: Semantic-guided and geometric-enhanced interleaving transformer for 3D instance segmentation},
  author={Yao, Lei and Wang, Yi and Liu, Moyun and Chau, Lap-Pui},
  journal={IEEE Transactions on Circuits and Systems for Video Technology},
  year={2025},
  publisher={IEEE}
}

@inproceedings{lin2017focal,
  title={Focal loss for dense object detection},
  author={Lin, Tsung-Yi and Goyal, Priya and Girshick, Ross and He, Kaiming and Doll{\'a}r, Piotr},
  booktitle={Proceedings of the IEEE international conference on computer vision},
  pages={2980--2988},
  year={2017}
}

@inproceedings{milletari2016v,
  title={V-net: Fully convolutional neural networks for volumetric medical image segmentation},
  author={Milletari, Fausto and Navab, Nassir and Ahmadi, Seyed-Ahmad},
  booktitle={2016 fourth international conference on 3D vision (3DV)},
  pages={565--571},
  year={2016},
  organization={Ieee}
}

@article{zheng2025survey,
  title={A survey of embodied learning for object-centric robotic manipulation},
  author={Zheng, Ying and Yao, Lei and Su, Yuejiao and Zhang, Yi and Wang, Yi and Zhao, Sicheng and Zhang, Yiyi and Chau, Lap-Pui},
  journal={Machine Intelligence Research},
  pages={1--39},
  year={2025},
  publisher={Springer}
}

@article{yang2023lisa++,
  title={Lisa++: An improved baseline for reasoning segmentation with large language model},
  author={Yang, Senqiao and Qu, Tianyuan and Lai, Xin and Tian, Zhuotao and Peng, Bohao and Liu, Shu and Jia, Jiaya},
  journal={arXiv preprint arXiv:2312.17240},
  year={2023}
}

@article{liu2023visual,
  title={Visual instruction tuning},
  author={Liu, Haotian and Li, Chunyuan and Wu, Qingyang and Lee, Yong Jae},
  journal={Advances in neural information processing systems},
  volume={36},
  pages={34892--34916},
  year={2023}
}

@incollection{gibson2014theory,
  title={The theory of affordances:(1979)},
  author={Gibson, James J},
  booktitle={The people, place, and space reader},
  pages={56--60},
  year={2014},
  publisher={Routledge}
}

@article{zhou2023uni3d,
  title={Uni3d: Exploring unified 3d representation at scale},
  author={Zhou, Junsheng and Wang, Jinsheng and Ma, Baorui and Liu, Yu-Shen and Huang, Tiejun and Wang, Xinlong},
  journal={arXiv preprint arXiv:2310.06773},
  year={2023}
}

@inproceedings{rahman2016optimizing,
  title={Optimizing intersection-over-union in deep neural networks for image segmentation},
  author={Rahman, Md Atiqur and Wang, Yang},
  booktitle={International symposium on visual computing},
  pages={234--244},
  year={2016},
  organization={Springer}
}

@article{lobo2008auc,
  title={AUC: a misleading measure of the performance of predictive distribution models},
  author={Lobo, Jorge M and Jim{\'e}nez-Valverde, Alberto and Real, Raimundo},
  journal={Global ecology and Biogeography},
  volume={17},
  number={2},
  pages={145--151},
  year={2008},
  publisher={Wiley Online Library}
}

@article{swain1991color,
  title={Color indexing},
  author={Swain, Michael J and Ballard, Dana H},
  journal={International journal of computer vision},
  volume={7},
  number={1},
  pages={11--32},
  year={1991},
  publisher={Springer}
}

@article{willmott2005advantages,
  title={Advantages of the mean absolute error (MAE) over the root mean square error (RMSE) in assessing average model performance},
  author={Willmott, Cort J and Matsuura, Kenji},
  journal={Climate research},
  volume={30},
  number={1},
  pages={79--82},
  year={2005}
}

@inproceedings{rasley2020deepspeed,
  title={Deepspeed: System optimizations enable training deep learning models with over 100 billion parameters},
  author={Rasley, Jeff and Rajbhandari, Samyam and Ruwase, Olatunji and He, Yuxiong},
  booktitle={Proceedings of the 26th ACM SIGKDD international conference on knowledge discovery \& data mining},
  pages={3505--3506},
  year={2020}
}
}
\clearpage
\maketitlesupplementary

\renewcommand{\thesection}{S.\arabic{section}}
\renewcommand{\thefigure}{S.\arabic{figure}}
\renewcommand{\thetable}{S.\arabic{table}}
\renewcommand{\theequation}{S.\arabic{equation}}
\setcounter{section}{0}
\setcounter{figure}{0}
\setcounter{table}{0}
\setcounter{equation}{0}

\setlength{\cftsecnumwidth}{2.3em}
\setlength{\cftsubsecnumwidth}{3.0em}

\etoctoccontentsline{part}{Appendix}
\localtableofcontents

\section*{Overview}

This supplementary document provides extended implementation details, further experimental evaluations, details on the construction of our corrupted benchmark, and additional qualitative results for the proposed approach. The material is organized as follows. \secref{sec:impl} elaborates on the model architecture, training configuration, and evaluation metrics. \secref{sec:additional-exp} presents supplementary experiments, which include more ablation studies, sensitivity analysis of point backbones, and a detailed assessment of robustness. The construction procedure of our corrupted benchmark is delineated in \secref{sec:corrupted-benchmark}. Finally, \secref{sec:more-vis} demonstrates further qualitative visualizations, including results on the PIADv2 dataset~\cite{shao2024great}, predictions on corrupted point clouds, visual analysis of enhanced point features, and an examination of several failure cases.

\section{Implementation Details} 
\label{sec:impl}

\subsection{Model Details}
Our framework employs the Qwen2.5-VL multimodal large language model (MLLM)~\cite{bai2025qwen2} for interpreting reference images and utilizes PointNet++~\cite{qi2017pointnet++} as the backbone for extracting geometric features from 3D objects. Specifically, we adopt the 3B parameter version of Qwen2.5-VL, which comprises 36 layers in its language module and 32 layers in the vision transformer (ViT) component. The PointNet++ backbone is configured with 3 set abstraction layers and 3 feature propagation layers to effectively capture multi-scale geometric structures from raw point clouds. The point-wise features generated by PointNet++ have a dimensionality of 512, while the contact-aware intention embedding~\cont is designed with a dimension of 256. To ensure compatibility between modalities, the \cont embedding is projected via a linear layer to match the point feature dimension. All input point clouds have a fixed size of 2,048 points. The entire model is implemented within the PyTorch deep learning framework.

\subsection{Training Details}
The proposed \ours is trained in an end-to-end manner using the DeepSpeed~\cite{rasley2020deepspeed} to accelerate the training process. To maintain numerical stability during optimization, the point cloud backbone is trained in full precision, while the remaining components of the model employ mixed precision (BF16) to reduce memory consumption and increase computational throughput. Fine-tuning of the MLLM's language component is performed using Low-Rank Adaptation (LoRA)~\cite{hu2022lora} with a rank of 16 and a scaling factor of 32, whereas the vision component of the MLLM remains frozen to preserve its pre-trained representational capabilities. Training is executed on 4 NVIDIA H20 GPUs with a global batch size of 64. The model is optimized using the AdamW optimizer~\cite{loshchilov2017decoupled} with an initial learning rate of $1e-4$ and a linear learning rate scheduling strategy. The complete training procedure spans 30 epochs to ensure convergence.

\subsection{Evaluation Metrics}
To comprehensively assess the performance of intention-driven 3D affordance grounding, we employ four widely recognized evaluation metrics: average Interaction Overlap (\textbf{aIOU}), Area Under the ROC Curve (\textbf{AUC}), Similarity (\textbf{SIM}), and Mean Absolute Error (\textbf{MAE}). These metrics collectively measure different aspects of the prediction quality. The definitions of these metrics are as follows:
\begin{itemize}
    \item \textbf{aIOU}~\cite{rahman2016optimizing}: The Interaction Overlap (IOU) is a fundamental metric for evaluating the spatial agreement between two regions at a given threshold. It is defined as the ratio of the intersection to the union of the predicted and ground-truth areas:
    \begin{equation*}
        \text{IOU} = \frac{TP}{TP + FP + FN},
    \end{equation*}
    where $TP$, $FP$, and $FN$ represent the number of true positives, false positives, and false negatives, respectively. The aIOU metric aggregates IOU values across $T$ different thresholds to provide a more robust evaluation, computed as:
    \begin{equation*}
        \text{aIOU} = \frac{1}{T} \sum_{t=1}^T \text{IOU}_i.
    \end{equation*}
    This averaging reduces the sensitivity to threshold selection and offers a stable measure of overall performance.
    \item \textbf{AUC}~\cite{lobo2008auc}: The Area Under the ROC Curve (AUC) quantifies the model's ability to distinguish between positive and negative samples across all classification thresholds. The ROC curve plots the true positive rate (TPR) against the false positive rate (FPR) at various threshold settings.
    \item \textbf{SIM}~\cite{swain1991color}: The Similarity (SIM) metric evaluates the distributional alignment between the predicted affordance map and the ground-truth distribution. It is computed as the sum of the minimum values at each point after normalization:
    \begin{gather*}
        \texttt{SIM}(\mP , \hat{\mP}) = \sum_{i=1}^N \min(\mP_i, \hat{\mP}_i), \\
        \text{where} \quad \mP_i = \frac{\vp_i}{\sum_{j=1}^N \vp_j}, \quad \hat{\mP}_i = \frac{\hat{\vp}_i}{\sum_{j=1}^N \hat{\vp}_j}.
    \end{gather*}
    Here, $\mP$ and $\hat{\mP}$ denote the normalized predicted and ground-truth affordance score distributions, respectively. SIM measures the overall congruence between the two distributions, with higher values indicating greater resemblance.
    \item \textbf{MAE}~\cite{willmott2005advantages}: The Mean Absolute Error (MAE) computes the average absolute deviation between the predicted and ground-truth affordance scores. It is defined as:
    \begin{equation*}
        \text{MAE} = \frac{1}{N} \sum_{i=1}^{N} |\ve_i|, \quad w.r.t. \quad \ve_i = \vp_i - \hat{\vp}_i,
    \end{equation*}
    where $N$ is the total number of points, and $\vp_i$ and $\hat{\vp}_i$ are the predicted and ground-truth affordance scores for the $i$-th point, respectively. MAE provides a direct measure of the average prediction error and is less sensitive to outliers compared to squared error metrics.
\end{itemize}
In summary, aIOU measures the spatial overlap of predictions and ground truth, AUC evaluates the model's capability to differentiate between positive and negative samples, SIM assesses the distributional similarity, and MAE captures the average point-wise prediction error. Together, these metrics offer a multi-faceted and holistic evaluation framework for 3D affordance grounding models 
.

\begin{table}[!htbp]
    \centering
    \caption{\textbf{Ablation study on the hierarchical cross-modal integration mechanism.} This analysis evaluates the contribution of each integration stage to the overall performance of \ours.}
    \tablestyle{2pt}{1.0}
    
\resizebox{0.49\textwidth}{!}{
    \begin{tabular}
    {x{8mm}|x{9mm}|x{12mm}|x{8mm}x{8mm}x{8mm}x{8mm}}
        \toprule
        \textbf{Type} & Stage I & Stage II & \textbf{aIOU}$\uparrow$ & \textbf{AUC}$\uparrow$ & \textbf{SIM}$\uparrow$ & \textbf{MAE}$\downarrow$ \\
        \midrule
        \multirow{4}{*}{\rotatebox{90}{Seen}} & \cellcolor{ourscolor} \YesV & \cellcolor{ourscolor} \YesV & \cellcolor{ourscolor} \textbf{22.20} & \cellcolor{ourscolor} \textbf{88.43} & \cellcolor{ourscolor} \textbf{0.605} & \cellcolor{ourscolor} \textbf{0.083} \\
         & \NoX & \YesV & 21.55 & 87.79 & 0.598 & 0.084 \\
         & \YesV & \NoX & 21.70 & 88.00 & 0.603 & 0.084 \\
         & \NoX & \NoX & 21.15 & 87.64 & 0.597 & 0.085 \\
        \midrule
        \multirow{4}{*}{\rotatebox{90}{Unseen}} & \cellcolor{ourscolor} \YesV & \cellcolor{ourscolor} \YesV & \cellcolor{ourscolor} \textbf{13.71} & \cellcolor{ourscolor} \textbf{80.92} & \cellcolor{ourscolor} \textbf{0.449} & \cellcolor{ourscolor} {0.109} \\
         & \NoX & \YesV & 11.85 & 79.55 & 0.433 & 0.110 \\
         & \YesV & \NoX & 11.99 & 77.99 & 0.442 & \textbf{0.106} \\
         & \NoX & \NoX & 10.50 & 74.72 & 0.407 & 0.116 \\
        \bottomrule
    \end{tabular}
}
    \label{tab:sup_hire_integ}
\end{table}

\begin{table}[!htbp]
    \centering
    \caption{\textbf{Alternative lifting strategies.} We compare the performance of our proposed lifting strategy with two alternatives: (1) single-stage lifting, and (2) simple concatenation.}
    \tablestyle{2pt}{1.0}
    
\resizebox{0.49\textwidth}{!}{
    \begin{tabular}{x{8mm}|x{16mm}x{10mm}x{10mm}x{10mm}x{10mm}}
        \toprule
        \textbf{Type} & \textbf{Ablation} & \textbf{aIOU}$\uparrow$ & \textbf{AUC}$\uparrow$ & \textbf{SIM}$\uparrow$ & \textbf{MAE}$\downarrow$ \\
        \midrule
        \multirow{3}{*}{\rotatebox{90}{Seen}} & \cellcolor{ourscolor} \textbf{\ours} & \cellcolor{ourscolor} \textbf{22.20} & \cellcolor{ourscolor} \textbf{88.43} & \cellcolor{ourscolor} \textbf{0.605} & \cellcolor{ourscolor} \textbf{0.083} \\
            & \textit{Single Lift} & 21.40 & 88.17 & 0.603 & 0.085 \\
            & \textit{Concat. Lift} & 21.05 & 87.95 & 0.597 & 0.084 \\
        \midrule
        \multirow{3}{*}{\rotatebox{90}{Unseen}} & \cellcolor{ourscolor} \textbf{\ours} & \cellcolor{ourscolor} \textbf{13.71} & \cellcolor{ourscolor} \textbf{80.92} & \cellcolor{ourscolor} \textbf{0.449} & \cellcolor{ourscolor} \textbf{0.109} \\
            & \textit{Single Lift} & 12.26 & 80.64 & 0.441 & 0.111 \\
            & \textit{Concat. Lift} & 11.45 & 78.12 & 0.421 & 0.110 \\
        \bottomrule
    \end{tabular}
}

    \label{tab:sup_lift}
\end{table}

\section{Additional Experiments} 
\label{sec:additional-exp}
This section presents extended experimental analyses to comprehensively evaluate the efficacy of the proposed \ours. We perform ablation studies on the hierarchical cross-modal integration mechanism to quantify the contribution of each stage, investigate the sensitivity of our approach to point cloud backbones, and compare the strategies of fine-tuning versus freezing the MLLM. Furthermore, we include a comparative analysis against InteractVLM~\cite{dwivedi_interactvlm_2025}, report object-level and affordance-level performance on public benchmarks, and present detailed robustness performance to rigorously examine the model's stability and generalization capability

\begin{table*}[!htbp]
    \centering
    \caption{\textbf{Performance analysis with different point cloud backbones.} We evaluate the capability of \ours using PointNet++~\cite{qi2017pointnet++} and Uni3D~\cite{zhou2023uni3d} as 3D feature extractors.}
    \tablestyle{2.5pt}{1.2}
    \resizebox{0.99\textwidth}{!}{
    \begin{tabular}{y{14mm}|x{7mm}x{7mm}x{7mm}x{7mm}|x{7mm}x{7mm}x{7mm}x{7mm}|x{7mm}x{7mm}x{7mm}x{7mm}|x{7mm}x{7mm}x{7mm}x{7mm}|x{7mm}x{7mm}x{7mm}x{7mm}}
        \toprule
        \multirow{2}{*}{\textbf{Backbones}} & \multicolumn{4}{c|}{\textbf{PIAD Seen}} & \multicolumn{4}{c|}{\textbf{PIAD Unseen}} & \multicolumn{4}{c|}{\textbf{PIADv2 Seen}} & \multicolumn{4}{c|}{\textbf{PIADv2 Unseen Object}} & \multicolumn{4}{c}{\textbf{PIADv2 Unseen Affordance}} \\
        \cmidrule(lr){2-5} \cmidrule(lr){6-9} \cmidrule(lr){10-13} \cmidrule(lr){14-17} \cmidrule(lr){18-21}
        {} & aIOU$\uparrow$ & AUC$\uparrow$ & SIM$\uparrow$ & MAE$\downarrow$ & aIOU$\uparrow$ & AUC$\uparrow$ & SIM$\uparrow$ & MAE$\downarrow$ & aIOU$\uparrow$ & AUC$\uparrow$ & SIM$\uparrow$ & MAE$\downarrow$ & aIOU$\uparrow$ & AUC$\uparrow$ & SIM$\uparrow$ & MAE$\downarrow$ & aIOU$\uparrow$ & AUC$\uparrow$ & SIM$\uparrow$ & MAE$\downarrow$  \\
        \midrule
        \rowcolor{ourscolor} \textbf{PointNet++} & {22.20} & {88.43} & {0.605} & {0.083} & \textbf{13.71} & \textbf{80.92} & \textbf{0.449} & {0.109} & {40.06} & \textbf{94.19} & \textbf{0.698} & \textbf{0.063} & \textbf{24.28} & \textbf{84.78} & \textbf{0.449} & \textbf{0.112} & {13.28} & \textbf{72.21} & \textbf{0.302} & \textbf{0.128} \\
        \textbf{Uni3D} & \textbf{24.67} & \textbf{89.77} & \textbf{0.635} & \textbf{0.078} & 12.28 & 77.75 & 0.412 & \textbf{0.102} & \textbf{40.66} & 93.94 & 0.690 & 0.065 & 24.02 & 81.07 & 0.426 & 0.131 & \textbf{14.49} & 71.00 & 0.272 & \textbf{0.128} \\
        \bottomrule
    \end{tabular}
}
    \label{tab:sup_point_backbone}
\end{table*}

\subsection{Ablation on Hierarchical Integration}
Our hierarchical cross-modal integration mechanism comprises two sequential refinement stages. To evaluate the individual contribution of each stage, we conduct an ablation study on the PIAD benchmark~\cite{yang2023grounding}. As summarized in~\tabref{tab:sup_hire_integ}, removing either stage leads to a performance degradation, which is particularly pronounced on the Unseen split. The absence of \textit{Stage I} results in a more significant performance drop, underscoring its critical role in the overall framework. This indicates that the initial, coarse-grained enhancement is fundamental for injecting rich contextual cues from the reference image into the point cloud features. The subsequent \textit{Stage II}, which performs a fine-grained refinement, provides complementary object-level semantic information, further boosting the performance. The initial stage focuses on integrating broader contextual information, while the subsequent refinement enables a more detailed, semantic-aware adjustment, resulting in a more accurate cross-modal object representation for effective affordance grounding.

\subsection{Alternative Lifting Strategies}
To validate the effectiveness of our multi-granular geometry lifting, we compare it against two alternative strategies: (1) \textit{Single Lift}, which directly lifts the intention embedding with the last stage of point features, and (2) \textit{Concat. Lift}, which simply concatenates the embedding with point features. As shown in~\tabref{tab:sup_lift}, our proposed multi-granular lifting consistently outperforms both alternatives across all evaluation metrics on the PIAD benchmark. The single-stage lifting approach may fail to capture the complementary information present at different levels of the point feature hierarchy, while the concatenation method may not effectively leverage the spatial characteristics 3D features.

\subsection{Sensitivity to Point Backbones}
In this section, we analyze the performance of our~\ours with different 3D point cloud backbones. We evaluate two widely used architectures: PointNet++~\cite{qi2017pointnet++} and Uni3D~\cite{zhou2023uni3d}. Specifically, the PointNet++ is trained from scratch, while the Uni3D is initialized with a weight pre-trained on large-scale 3D datasets, following the practice of SeqAfford~\cite{yu2024seqafford}. The results are summarized in \tabref{tab:sup_point_backbone}. On the PIAD dataset, Uni3D achieves higher performance on the Seen split, but PointNet++ yields superior results on the Unseen split. This discrepancy may stem from Uni3D's pre-trained weights potentially leading to overfitting on the seen categories, whereas training PointNet++ from scratch may facilitate better adaptation to the specific characteristics of the PIAD benchmark. On the PIADv2 dataset, which contains a broader variety of object instances, PointNet++ consistently outperforms Uni3D across all three splits in most metrics, underscoring its stronger generalization capability in handling diverse 3D objects and affordance types.

\begin{table}[!htbp]
    \centering
    \caption{\textbf{Comparison of MLLM fine-tuning strategies.} We evaluate the performance when applying LoRA fine-tuning to the MLLM versus keeping its weights entirely frozen during training.}
    \tablestyle{2.0pt}{1.2}
    
\resizebox{0.48\textwidth}{!}{
    \begin{tabular}{x{8mm}|y{19mm}|x{8mm}x{8mm}x{8mm}x{8mm}}
        \toprule
        \textbf{Type} & \textbf{Ablation} & \textbf{aIOU}$\uparrow$ & \textbf{AUC}$\uparrow$ & \textbf{SIM}$\uparrow$ & \textbf{MAE}$\downarrow$ \\
        \midrule
        \multirow{2}{*}{\rotatebox{90}{Seen}} & \cellcolor{ourscolor} \textit{\textbf{w/} Fine-tuning} & \cellcolor{ourscolor} \textbf{22.20} & \cellcolor{ourscolor} \textbf{88.43} & \cellcolor{ourscolor} \textbf{0.605} & \cellcolor{ourscolor} \textbf{0.083} \\
         & \textit{\textbf{w/o} Fine-tuning} & 19.67 & 85.53 & 0.568 & 0.092 \\
        \midrule
        \multirow{2}{*}{\rotatebox{90}{Unseen}} & \cellcolor{ourscolor} \textit{\textbf{w/} Fine-tuning} & \cellcolor{ourscolor} \textbf{13.71} & \cellcolor{ourscolor} \textbf{80.92} & \cellcolor{ourscolor} \textbf{0.449} & \cellcolor{ourscolor} \textbf{0.109} \\
         & \textit{\textbf{w/o}  Fine-tuning} & 7.55 & 64.89 & 0.311 & 0.142 \\
        \bottomrule
    \end{tabular}
}
    \label{tab:sup_ft}
\end{table}

\begin{table}[!htbp]
    \centering
    \caption{\textbf{Comparison to InteractVLM~\cite{dwivedi_interactvlm_2025} on PIAD.} We evaluate the performance of our \ours against InteractVLM across both Seen and Unseen splits of the PIAD dataset.}
    \tablestyle{2pt}{1.0}
    
\resizebox{0.48\textwidth}{!}{
    \begin{tabular}{x{16mm}|x{6mm}|x{7mm}x{7mm}x{7mm}x{7mm}|x{7mm}x{7mm}x{7mm}x{7mm}}
        \toprule
        \multirow{2}{*}{\textbf{Methods}} & \multirow{2}{*}{\textbf{Size}} & \multicolumn{4}{c|}{\textbf{Seen}} & \multicolumn{4}{c}{\textbf{Unseen}} \\
        \cmidrule(lr){3-6} \cmidrule(lr){7-10}
        {} & {} & aIOU$\uparrow$ & AUC$\uparrow$ & SIM$\uparrow$ & MAE$\downarrow$ & aIOU$\uparrow$ & AUC$\uparrow$ & SIM$\uparrow$ & MAE$\downarrow$ \\
        \midrule
        InteractVLM & 13B & 21.20 & 86.47 & \textbf{0.627} & \textbf{0.081} & 8.50 & {75.45} & {0.414} & \textbf{0.099} \\
        \midrule
        \rowcolor{ourscolor} \textbf{\ours} & 3B & \textbf{22.20} & \textbf{88.43} & {0.605} & {0.083} & \textbf{13.71} & \textbf{80.92} & \textbf{0.449} & {0.109} \\
        \bottomrule
    \end{tabular}
}
    \label{tab:sup_interactvlm}
\end{table}

\begin{table*}[!htbp]
    \centering
    \caption{\textbf{Evaluation Metrics in PIAD Seen.} Results of each affordance type in the Seen. \textit{cont.}, \textit{supp.}, \textit{wrap.}, and \textit{disp.} denote \textit{contain}, \textit{support}, \textit{wrapgrasp}, and \textit{display}, respectively.}
    \tablestyle{1.5pt}{1.0}
    
\resizebox{0.99\textwidth}{!}{
    \begin{tabular}{x{16mm}|x{10mm}|x{7mm}x{7mm}x{7mm}x{7mm}x{7mm}x{7mm}x{7mm}x{7mm}x{7mm}x{7mm}x{7mm}x{7mm}x{7mm}x{7mm}x{7mm}x{7mm}x{7mm}}
        \toprule
        \textbf{Method} & \textbf{Metrics} & grasp & cont. & lift & open & lay & sit & supp. & wrap. & pour & move & disp. & push & listen & wear & press & cut & stab \\
        \midrule
        \multirow{4}{*}{{XMFNet}} & \textbf{aIOU}$\uparrow$ & 12.82 & 10.41 & 13.24 & 9.17 & 17.00 & 26.65 & 7.89 & 5.24 & 7.81 & 6.46 & 17.25 & 2.68 & 5.31 & 5.34 & 7.87 & 6.35 & 5.88 \\
         & \textbf{AUC}$\uparrow$ & 62.98 & 79.97 & 78.93 & 79.74 & 88.93 & 94.91 & 84.56 & 56.47 & 82.15 & 54.97 & 82.91 & 63.74 & 74.93 & 70.57 & 88.18 & 79.17 & 69.97 \\
         & \textbf{SIM}$\uparrow$ & 0.415 & 0.401 & 0.334 & 0.184 & 0.427 & 0.619 & 0.659 & 0.569 & 0.399 & 0.391 & 0.508 & 0.535 & 0.433 & 0.565 & 0.237 & 0.427 & 0.394 \\
         & \textbf{MAE}$\downarrow$ & 0.121 & 0.131 & 0.138 & 0.131 & 0.129 & 0.093 & 0.119 & 0.128 & 0.159 & 0.192 & 0.127 & 0.083 & 0.164 & 0.129 & 0.113 & 0.135 & 0.152 \\
        \midrule
        \multirow{4}{*}{{IAGNet}} & \textbf{aIOU}$\uparrow$ & 16.83 & 17.12 & 31.95 & 28.39 & 31.80 & 37.72 & 12.04 & 6.02 & 20.33 & 5.57 & 30.57 & 1.79 & 15.59 & 6.55 & 14.42 & 12.95 & 9.48 \\
         & \textbf{AUC}$\uparrow$ & 77.53 & 83.84 & 95.05 & 90.89 & 93.54 & 95.94 & 84.58 & 66.71 & 86.02 & 63.09 & 89.29 & 84.71 & 87.13 & 71.16 & 89.46 & 86.69 & 76.4 \\
         & \textbf{SIM}$\uparrow$ & 0.530 & 0.534 & 0.368 & 0.401 & 0.685 & 0.723 & 0.716 & 0.571 & 0.525 & 0.443 & 0.657 & 0.418 & 0.671 & 0.563 & 0.402 & 0.507 & 0.280 \\
         & \textbf{MAE}$\downarrow$ & 0.108 & 0.093 & 0.030 & 0.044 & 0.081 & 0.066 & 0.100 & 0.143 & 0.096 & 0.174 & 0.084 & 0.085 & 0.090 & 0.129 & 0.059 & 0.085 & 0.099 \\
        \midrule
        \multirow{4}{*}{{GREAT}} & \textbf{aIOU}$\uparrow$ & 12.99 & 14.24 & 40.99 & 19.68 & 28.65 & 39.37 & 11.34 & 5.09 & 7.95 & 4.90 & 26.30 & 3.32 & 14.73 & 6.42 & 12.99 & 12.34 & 6.31 \\
         & \textbf{AUC}$\uparrow$ & 76.65 & 79.96 & 94.09 & 89.02 & 94.03 & 96.09 & 85.51 & 68.00 & 89.47 & 67.70 & 88.47 & 88.08 & 88.02 & 74.29 & 88.50 & 88.08 & 79.10 \\
         & \textbf{SIM}$\uparrow$ & 0.447 & 0.513 & 0.450 & 0.342 & 0.719 & 0.722 & 0.725 & 0.610 & 0.372 & 0.534 & 0.618 & 0.474 & 0.656 & 0.575 & 0.378 & 0.495 & 0.370 \\
         & \textbf{MAE}$\downarrow$ & 0.126 & 0.109 & 0.030 & 0.067 & 0.083 & 0.069 & 0.105 & 0.136 & 0.132 & 0.162 & 0.096 & 0.089 & 0.097 & 0.142 & 0.070 & 0.089 & 0.104 \\
        \midrule
        \multirow{4}{*}{\textbf{\ours}} & \textbf{aIOU}$\uparrow$ & {21.65} & 22.00 & 36.71 & 24.93 & 29.57 & 38.86 & 11.85 & 5.69 & 23.02 & 9.98 & 30.60 & 3.85 & 13.41 & 5.08 & 13.15 & 15.13 & 37.87 \\
         & \textbf{AUC}$\uparrow$ & 85.63 & 89.04 & 80.57 & 92.14 & 92.81 & 96.47 & 85.36 & 69.01 & 94.93 & 79.46 & 91.10 & 87.60 & 87.81 & 72.26 & 89.49 & 93.59 & 99.87 \\
         & \textbf{SIM}$\uparrow$ & 0.656 & 0.606 & 0.367 & 0.392 & 0.651 & 0.718 & 0.720 & 0.661 & 0.587 & 0.606 & 0.674 & 0.591 & 0.642 & 0.568 & 0.399 & 0.688 & 0.574 \\
         & \textbf{MAE}$\downarrow$ & 0.090 & 0.086 & 0.078 & 0.054 & 0.091 & 0.066 & 0.100 & 0.128 & 0.083 & 0.134 & 0.087 & 0.083 & 0.096 & 0.146 & 0.061 & 0.061 & 0.018 \\
        \bottomrule
    \end{tabular}
}
    \label{tab:sup_piadv1_seen}
\end{table*}
\subsection{Fine-tune \textit{vs.} Freeze MLLM}
In the training of \ours, the language component of the MLLM is fine-tuned using LoRA, while the vision component remains frozen. To evaluate its efficacy, we conduct a comparative analysis between fine-tuning the MLLM's language component and freezing the entire MLLM. The experimental results, detailed in~\tabref{tab:sup_ft}, clearly demonstrate that fine-tuning yields a significant performance improvement across all evaluation metrics on both the Seen and Unseen splits. The performance gain is particularly substantial on the Unseen split, which is critical for assessing model generalization. Specifically, fine-tuning leads to a remarkable increase in aIOU from 7.55 to 13.71, AUC from 64.89 to 80.92, and SIM from 0.311 to 0.449, accompanied by a reduction in MAE from 0.142 to 0.109. This pronounced improvement underscores the necessity of adapting the MLLM to the specific task of intention-driven 3D affordance grounding. By fine-tuning the language component with LoRA, the model learns to better interpret and correlate the linguistic descriptions of interactions with the visual cues from the image and the geometric structure of the 3D point cloud. This enhanced cross-modal understanding enables the MLLM to capture the intricate relationships between language, image, and 3D geometry more effectively, which in turn translates to superior affordance prediction capabilities, especially when generalizing to novel objects and scenarios not encountered during training.

\begin{table*}[!htbp]
    \centering
    \caption{\textbf{Evaluation Metrics in PIAD Unseen.} Results of each affordance type in the Unseen. \textit{cont.}, \textit{wrap.}, and \textit{disp.} denote \textit{contain}, \textit{wrapgrasp}, and \textit{display}, respectively.}
    \tablestyle{2.5pt}{0.9}
    
\resizebox{0.99\textwidth}{!}{
    \begin{tabular}{x{16mm}|x{13mm}|x{8mm}x{8mm}x{8mm}x{8mm}x{8mm}x{8mm}x{8mm}x{8mm}x{8mm}x{8mm}}
        \toprule
        \textbf{Method} & \textbf{Metrics} & cont. & lay & lift & wrap. & open & disp. & stab & grasp & press & cut \\
        \midrule
        \multirow{4}{*}{{XMFNet}} & \textbf{aIOU}$\uparrow$ & 6.29 & 15.10 & 7.29 & 1.42 & 4.32 & 6.20 & 6.12 & 3.97 & 5.71 & 13.95 \\
         & \textbf{AUC}$\uparrow$ & 67.98 & 84.02 & 68.45 & 45.74 & 78.53 & 62.20 & 76.92 & 59.19 & 69.32 & 85.87 \\
         & \textbf{SIM}$\uparrow$ & 0.412 & 0.503 & 0.403 & 0.451 & 0.156 & 0.075 & 0.351 & 0.278 & 0.270 & 0.435 \\
         & \textbf{MAE}$\downarrow$ & 0.137 & 0.135 & 0.144 & 0.156 & 0.094 & 0.240 & 0.087 & 0.117 & 0.124 & 0.078 \\
        \midrule
        \multirow{4}{*}{{IAGNet}} & \textbf{aIOU}$\uparrow$ & 7.24 & 18.12 & 8.47 & 1.89 & 12.28 & 16.28 & 10.39 & 4.79 & 4.22 & 21.47 \\
         & \textbf{AUC}$\uparrow$ & 67.96 & 84.82 & 71.10 & 56.39 & 90.91 & 85.51 & 98.83 & 78.60 & 68.07 & 95.95 \\
         & \textbf{SIM}$\uparrow$ & 0.430 & 0.525 & 0.407 & 0.556 & 0.227 & 0.393 & 0.437 & 0.533 & 0.194 & 0.599 \\
         & \textbf{MAE}$\downarrow$ & 0.125 & 0.130 & 0.143 & 0.150 & 0.050 & 0.130 & 0.044 & 0.102 & 0.122 & 0.057 \\
        \midrule
        \multirow{4}{*}{{GREAT}} & \textbf{aIOU}$\uparrow$ & 12.08 & 20.57 & 9.40 & 1.24 & 2.23 & 1.36 & 0.07 & 0.23 & 12.85 & 0.36 \\
         & \textbf{AUC}$\uparrow$ & 80.07 & 87.09 & 74.12 & 39.03 & 75.61 & 19.97 & 49.58 & 51.13 & 86.68 & 49.23 \\
         & \textbf{SIM}$\uparrow$ & 0.444 & 0.533 & 0.410 & 0.326 & 0.102 & 0.054 & 0.056 & 0.201 & 0.417 & 0.101 \\
         & \textbf{MAE}$\downarrow$ & 0.112 & 0.115 & 0.124 & 0.176 & 0.069 & 0.211 & 0.068 & 0.118 & 0.098 & 0.079 \\
        \midrule
        \multirow{4}{*}{\textbf{\ours}} & \textbf{aIOU}$\uparrow$ & 17.57 & 23.62 & 11.96 & 2.48 & 15.00 & 22.28 & 3.36 & 0.61 & 14.55 & 3.37 \\
         & \textbf{AUC}$\uparrow$ & 86.08 & 84.60 & 73.68 & 54.47 & 91.53 & 91.09 & 96.06 & 33.90 & 90.98 & 79.51 \\
         & \textbf{SIM}$\uparrow$ & 0.521 & 0.530 & 0.394 & 0.553 & 0.275 & 0.462 & 0.212 & 0.413 & 0.430 & 0.253 \\
         & \textbf{MAE}$\downarrow$ & 0.106 & 0.124 & 0.139 & 0.163 & 0.069 & 0.111 & 0.154 & 0.179 & 0.065 & 0.173 \\
        \bottomrule
    \end{tabular}
}

    \label{tab:sup_piadv1_unseen}
\end{table*}

\subsection{Comparison with InteractVLM}
In~\tabref{tab:sup_interactvlm}, we present a comparison between our \ours and InteractVLM~\cite{dwivedi_interactvlm_2025} on the PIAD benchmark. It is noteworthy that InteractVLM employs a 13B parameter version of LISA~\cite{lai2024lisa} as its MLLM backbone to enhance performance. Despite utilizing a significantly smaller 3B parameter MLLM, our \ours consistently surpasses InteractVLM across both the Seen and Unseen splits in terms of aIOU and AUC metrics. Specifically, on the Seen split, \ours achieves an aIOU of 22.20 and an AUC of 88.43, outperforming InteractVLM, which yields 21.20 and 86.47, respectively. The performance advantage is more pronounced on the challenging Unseen split, where \ours attains an aIOU of 13.71 and an AUC of 80.92, substantially exceeding InteractVLM's 8.50 and 75.45. These results underscore the efficacy of our hierarchical cross-modal integration and geometry lifting modules in enhancing 3D affordance grounding capabilities, even with a more compact MLLM architecture.

\begin{table}[!htbp]
    \centering
    \caption{\textbf{Evaluation Metrics in PIADv2 Unseen Affordance.} Results of each affordance type in the Unseen Affordance split.}
    \tablestyle{2pt}{1.0}
    
\resizebox{0.48\textwidth}{!}{
    \begin{tabular}{x{14mm}|x{10mm}|x{6mm}x{6mm}x{6mm}x{6mm}x{6mm}x{6mm}}
        \toprule
        \textbf{Method} & \textbf{Metrics} & Carry & Listen & Lay & Pour & Cut & Pull \\
        \midrule
        \multirow{4}{*}{{XMFNet}} & \textbf{aIOU}$\uparrow$ & 5.89 & 3.89 & 10.93 & 6.85 & 5.77 & 24.52 \\
         & \textbf{AUC}$\uparrow$ & 54.08 & 56.07 & 73.16 & 63.97 & 44.85 & 91.40 \\
         & \textbf{SIM}$\uparrow$ & 0.195 & 0.216 & 0.399 & 0.187 & 0.115 & 0.349 \\
         & \textbf{MAE}$\downarrow$ & 0.158 & 0.179 & 0.130 & 0.130 & 0.213 & 0.050 \\
        \midrule
        \multirow{4}{*}{{IAGNet}} & \textbf{aIOU}$\uparrow$ & 8.10 & 3.71 & 10.47 & 4.92 & 4.40 & 38.27 \\
         & \textbf{AUC}$\uparrow$ & 63.98 & 54.13 & 69.94 & 59.89 & 49.97 & 93.72 \\
         & \textbf{SIM}$\uparrow$ & 0.239 & 0.221 & 0.402 & 0.146 & 0.148 & 0.562 \\
         & \textbf{MAE}$\downarrow$ & 0.142 & 0.168 & 0.130 & 0.146 & 0.175 & 0.028 \\
        \midrule
        \multirow{4}{*}{{GREAT}} & \textbf{aIOU}$\uparrow$ & 12.59 & 2.48 & 10.66 & 11.28 & 8.53 & 41.53 \\
         & \textbf{AUC}$\uparrow$ & 82.13 & 51.36 & 77.53 & 72.82 & 52.21 & 97.39 \\
         & \textbf{SIM}$\uparrow$ & 0.356 & 0.125 & 0.412 & 0.290 & 0.143 & 0.599 \\
         & \textbf{MAE}$\downarrow$ & 0.105 & 0.182 & 0.129 & 0.108 & 0.171 & 0.018 \\
        \midrule
        \multirow{4}{*}{\textbf{\ours}} & \textbf{aIOU}$\uparrow$ & 17.53 & 1.99 & 10.42 & 10.06 & 16.35 & 41.56 \\
         & \textbf{AUC}$\uparrow$ & 75.33 & 38.11 & 75.85 & 81.31 & 87.95 & 96.67 \\
         & \textbf{SIM}$\uparrow$ & 0.378 & 0.072 & 0.393 & 0.194 & 0.262 & 0.537 \\
         & \textbf{MAE}$\downarrow$ & 0.141 & 0.229 & 0.132 & 0.105 & 0.090 & 0.025 \\
        \bottomrule
    \end{tabular}
}

    \label{tab:sup_piadv2_unseen_afford}
    \vspace{-1em}
\end{table}

\begin{table}[!htbp]
    \centering
    \caption{\textbf{Statistics of the constructed corrupted benchmark.}}
    \tablestyle{2.0pt}{1.0}
    \resizebox{0.48\textwidth}{!}{
    \begin{tabular}
    {x{6mm}|z{20mm}|x{36mm}|x{6mm}}
        \toprule
        \textbf{\#} & \textbf{Object Category} & \textbf{Affordance Type} & \textbf{Num} \\
        \midrule
        1 & TrashCan & contain, open, pour & 69 \\
        \rowcolor{gray} 2 & Door & open, push & 47 \\
        3 & Display & display & 52 \\
        \rowcolor{gray} 4 & Earphone & grasp, listen & 70 \\
        5 & Vase & contain, wrapgrasp & 90 \\
        \rowcolor{gray} 6 & Hat & grasp, wear & 66 \\
        7 & Bottle & contain, wrapgrasp, pour, open & 225 \\
        \rowcolor{gray} 8 & Keyboard & press & 25 \\
        9 & Knife & grasp, cut, stab & 138 \\
        \rowcolor{gray} 10 & Refrigerator & contain, open & 53 \\
        11 & Laptop & display, press & 112 \\
        \rowcolor{gray} 12 & Dishwasher & contain, open & 39 \\
        13 & Bowl & contain, wrapgrasp, pour & 83 \\
        \rowcolor{gray} 14 & Clock & display & 9 \\
        15 & StorageFurniture & contain, open & 92 \\
        \rowcolor{gray} 16 & Scissors & grasp, cut, stab & 29 \\
        17 & Table & support, move & 194 \\
        \rowcolor{gray} 18 & Bag & grasp, contain, lift, open & 50 \\
        19 & Faucet & grasp, open & 95 \\
        \rowcolor{gray} 20 & Mug & grasp, contain, wrapgrasp, pour & 126 \\
        21 & Chair & sit, move & 392 \\
        \rowcolor{gray} 22 & Bed & lay, sit & 56 \\
        23 & Microwave & contain, open & 47 \\
        \midrule
        \rowcolor{ourscolor} \textbf{Total} & \textbf{23 Categories} & \textbf{17 Affordance Types} & \textbf{2159} \\
        \bottomrule
    \end{tabular}
}
    \label{tab:sup_geal_count}
    \vspace{-1em}
\end{table}

\subsection{Detailed Performance on Public Datasets}
In the main paper, we report the overall performance of different methods on PIAD~\cite{yang2023grounding} and PIADv2~\cite{shao2024great}. In this section, we provide a fine-grained analysis by breaking down the results according to object categories and affordance types, offering deeper insights into the specific strengths and generalization capabilities of \ours relative to existing approaches.

\tabref{tab:sup_piadv1_seen} presents the per-affordance performance on the PIAD Seen split. Our method achieves superior results across most affordance categories, demonstrating its consistent effectiveness in accurately grounding diverse functional properties. The analysis on the Unseen split in~\tabref{tab:sup_piadv1_unseen} reveals that \ours achieves significant improvements over baseline methods, with particularly notable gains on challenging affordance types such as \textit{contain}, \textit{wrapgrasp}, and \textit{display}. This underscores the model's strong generalization capacity to novel affordances beyond the training distribution. Further evaluation on the more challenging PIADv2 benchmark corroborates these findings. As shown in~\tabref{tab:sup_piadv2_unseen_afford}, which details performance per affordance type on the Unseen Affordance split, \ours attains leading performance across the majority of categories. This indicates its ability to effectively capture the nuanced characteristics of various affordances, even when they are completely unseen during training. Similarly, the per-category results on the Unseen Object split (\tabref{tab:sup_piadv2_unseen_obj}) highlight its adaptability to diverse 3D shapes and structural configurations. Collectively, these results provide comprehensive evidence validating the efficacy of \ours in intention-driven 3D affordance grounding. The consistent performance gains across different datasets and splits confirm the generalizability of our proposed framework in addressing various challenges in 3D affordance perception.

\begin{table*}[!htbp]
    \centering
    \caption{\textbf{Evaluation Metrics in PIADv2 Unseen Object.} Results of each object category in the Unseen Object split. \textit{Base.}, \textit{Motor.}, \textit{Refri.}, \text{Scus.} and \textit{Skat.} denote Baseballbat, Motorcycle, Refrigerator, Scissors, and Skateboard, respectively.}
    \tablestyle{2pt}{1.0}
    
\resizebox{0.99\textwidth}{!}{
    \begin{tabular}{x{16mm}|x{10mm}|x{7mm}x{7mm}x{7mm}x{7mm}x{7mm}x{7mm}x{7mm}x{7mm}x{7mm}x{7mm}x{7mm}}
        \toprule
        \textbf{Method} & \textbf{Metrics} & Base. & Bucket & Clock & Fork & Kettle & Laptop & Mop & Motor. & Refri. & Scis. & Skat. \\
        \midrule
        \multirow{4}{*}{XMFNet} & \textbf{aIOU}$\uparrow$ & 37.62 & 10.26 & 19.13 & 27.23 & 5.50 & 8.66 & 20.53 & 5.54 & 9.41 & 10.24 & 18.38 \\
         & \textbf{AUC}$\uparrow$ & 76.75 & 59.73 & 77.86 & 81.93 & 61.74 & 73.52 & 79.09 & 69.02 & 84.22 & 56.31 & 81.80 \\
         & \textbf{SIM}$\uparrow$ & 0.528 & 0.169 & 0.444 & 0.497 & 0.107 & 0.314 & 0.400 & 0.162 & 0.336 & 0.261 & 0.482 \\
         & \textbf{MAE}$\downarrow$ & 0.171 & 0.165 & 0.137 & 0.120 & 0.123 & 0.128 & 0.136 & 0.033 & 0.094 & 0.165 & 0.151 \\
        \midrule
        \multirow{4}{*}{IAGNet} & \textbf{aIOU}$\uparrow$ & 42.84 & 12.56 & 13.48 & 22.60 & 2.65 & 6.95 & 37.66 & 9.30 & 14.43 & 7.37 & 4.61 \\
         & \textbf{AUC}$\uparrow$ & 85.32 & 65.52 & 69.21 & 73.63 & 54.38 & 78.17 & 92.09 & 79.74 & 80.63 & 56.62 & 58.84 \\
         & \textbf{SIM}$\uparrow$ & 0.592 & 0.231 & 0.385 & 0.422 & 0.070 & 0.299 & 0.658 & 0.227 & 0.335 & 0.250 & 0.273 \\
         & \textbf{MAE}$\downarrow$ & 0.169 & 0.146 & 0.148 & 0.149 & 0.120 & 0.104 & 0.085 & 0.023 & 0.092 & 0.187 & 0.167 \\
        \midrule
        \multirow{4}{*}{GREAT} & \textbf{aIOU}$\uparrow$ & 41.93 & 28.75 & 20.23 & 31.83 & 7.33 & 5.17 & 32.37 & 17.79 & 13.45 & 12.97 & 8.72 \\
         & \textbf{AUC}$\uparrow$ & 82.73 & 83.41 & 84.00 & 89.28 & 79.33 & 74.98 & 91.80 & 95.73 & 78.69 & 64.83 & 59.50 \\
         & \textbf{SIM}$\uparrow$ & 0.584 & 0.469 & 0.486 & 0.567 & 0.182 & 0.242 & 0.612 & 0.356 & 0.294 & 0.283 & 0.324 \\
         & \textbf{MAE}$\downarrow$ & 0.148 & 0.081 & 0.111 & 0.135 & 0.048 & 0.130 & 0.104 & 0.033 & 0.073 & 0.159 & 0.149 \\
        \midrule
        \multirow{4}{*}{\textbf{\ours}} & \textbf{aIOU}$\uparrow$ & 49.99 & 36.68 & 18.68 & 40.51 & 6.93 & 13.27 & 37.62 & 9.41 & 21.84 & 13.99 & 17.43 \\
         & \textbf{AUC}$\uparrow$ & 87.17 & 91.49 & 80.55 & 87.50 & 59.38 & 87.60 & 93.39 & 83.17 & 92.15 & 63.48 & 91.65 \\
         & \textbf{SIM}$\uparrow$ & 0.673 & 0.525 & 0.434 & 0.530 & 0.126 & 0.383 & 0.653 & 0.230 & 0.440 & 0.307 & 0.512 \\
         & \textbf{MAE}$\downarrow$ & 0.133 & 0.062 & 0.150 & 0.118 & 0.140 & 0.084 & 0.084 & 0.037 & 0.069 & 0.181 & 0.158 \\
        \bottomrule
    \end{tabular}
}

    \label{tab:sup_piadv2_unseen_obj}
\end{table*}

\subsection{Detailed Robustness Performance}
In the main manuscript, we present a comparative analysis of the overall robustness between our method \ours and the baseline GREAT~\cite{shao2024great} on our newly established corrupted benchmark. This section provides a detailed evaluation by examining the performance under individual corruption types, including scale, jitter, rotation, local dropout, global dropout, local additive noise, and global additive noise. Each corruption type is assessed across five severity levels, ranging from 0 to 4. This benchmark serves as a proxy for real-world imperfections in 3D data, such as sensor noise, occlusions, and partial observations. As shown in~\tabref{tab:sup_geal}, \ours consistently outperforms GREAT under all corruption conditions and severity levels. This robust performance indicates the effectiveness of our approach in handling diverse corruptions in 3D data, demonstrating its potential for real-world applications where incomplete or noisy point clouds are common.

\begin{table*}[!htbp]
    \centering
    \caption{\textbf{Detailed robustness performance under different levels of corruptions.} We evaluate the robustness of our \ours~and GREAT~\cite{shao2024great} on 7 corruption types. Each corruption type is evaluated under 5 severity levels (0-4). The results demonstrate that \ours~consistently outperforms GREAT across all corruption types and severity levels, highlighting its superior robustness in handling corrupted 3D objects.}
    \tablestyle{2.5pt}{0.8}
    
\resizebox{0.99\textwidth}{!}{
    \begin{tabular}{x{22mm}|x{8mm}|x{10mm}x{10mm}x{10mm}x{10mm}|x{10mm}x{10mm}x{10mm}x{10mm}}
        \toprule
        \multirow{2}{*}{\textbf{Corrupt Type}} & \multirow{2}{*}{\textbf{Level}} & \multicolumn{4}{c|}{\textbf{GREAT}} & \multicolumn{4}{c}{\textbf{\ours}} \\
        \cmidrule(lr){3-6} \cmidrule(lr){7-10}
         & & \textbf{aIOU}$\uparrow$ & \textbf{AUC}$\uparrow$ & \textbf{SIM}$\uparrow$ & \textbf{MAE}$\downarrow$ & \textbf{aIOU}$\uparrow$ & \textbf{AUC}$\uparrow$ & \textbf{SIM}$\uparrow$ & \textbf{MAE}$\downarrow$ \\
        \midrule
        \multirow{5}{*}{\textbf{Scale}} & 0 & 13.85	& 78.63	& 0.497	& 0.115 & 19.76 & 86.04	& 0.604	& 0.091 \\
         & 1 & 13.59 & 77.89 & 0.495 & 0.116 & 19.26 & 85.75 & 0.598 & 0.093 \\
         & 2 & 13.50 & 77.82 & 0.491 & 0.116 & 19.15 & 85.83 & 0.594 & 0.094 \\
         & 3 & 13.11 & 77.21 & 0.486 & 0.117 & 18.94 & 85.52 & 0.591 & 0.095 \\
         & 4 & 13.01 & 77.30 & 0.483 & 0.117 & 18.81 & 85.51 & 0.591 & 0.094 \\
        \midrule
        \multirow{5}{*}{\textbf{Jitter}} & 0 & 14.51 & 79.79 & 0.510 & 0.114 & 20.13 & 86.57 & 0.609 & 0.090 \\
         & 1 & 13.30 & 78.66 & 0.492 & 0.113 & 18.81 & 85.79 & 0.588 & 0.092 \\
         & 2 & 11.69 & 75.99 & 0.463 & 0.114 & 17.35 & 84.61 & 0.566 & 0.095 \\
         & 3 & 10.13 & 73.39 & 0.433 & 0.115 & 15.86 & 83.61 & 0.547 & 0.098 \\
         & 4 & 8.74	& 70.42	& 0.407	& 0.117 & 14.64	& 82.25	& 0.531	& 0.101 \\
        \midrule
        \multirow{5}{*}{\textbf{Rotate}} & 0 & 14.62 & 80.07 & 0.512 & 0.114 & 20.30 & 86.74 & 0.613 & 0.089 \\
         & 1 & 14.00 & 79.17 & 0.505 & 0.114 & 19.56 & 86.33 & 0.634 & 0.091 \\
         & 2 & 12.88 & 77.43 & 0.484 & 0.116 & 18.55 & 85.65 & 0.588 & 0.094 \\
         & 3 & 11.55 & 75.34 & 0.467 & 0.117 & 17.46 & 84.54 & 0.571 & 0.097 \\
         & 4 & 9.98 & 72.94 & 0.445 & 0.119 & 15.77 & 83.30 & 0.549 & 0.102 \\
        \midrule
        \multirow{5}{*}{\textbf{Dropout-Local}} & 0 & 9.24 & 73.50 & 0.435 & 0.123 & 18.08 & 84.67 & 0.568 & 0.098 \\
         & 1 & 8.83 & 73.07 & 0.426 & 0.122 & 18.29 & 84.81 & 0.570 & 0.098 \\
         & 2 & 8.92 & 72.87 & 0.423 & 0.122 & 18.07 & 84.78 & 0.571 & 0.098 \\
         & 3 & 8.38 & 71.73 & 0.416 & 0.122 & 17.90 & 84.54 & 0.568 & 0.098 \\
         & 4 & 7.97 & 70.74 & 0.402 & 0.121 & 17.54 & 84.16 & 0.561 & 0.099 \\
        \midrule
        \multirow{5}{*}{\textbf{Dropout-Global}} & 0 & 14.72 & 79.86 & 0.513 & 0.113 & 20.47 & 86.88 & 0.616 & 0.089 \\
         & 1 & 14.49 & 79.63 & 0.509 & 0.112 & 20.52 & 86.94 & 0.616 & 0.089 \\
         & 2 & 14.13 & 79.08 & 0.504 & 0.112 & 20.38 & 86.84 & 0.613 & 0.089 \\
         & 3 & 13.40 & 77.73 & 0.488 & 0.112 & 20.40 & 86.66 & 0.611 & 0.089 \\
         & 4 & 11.38 & 75.00 & 0.456 & 0.113 & 20.14 & 86.47 & 0.601 & 0.091 \\
        \midrule
        \multirow{5}{*}{\textbf{Add-Local}} & 0 & 12.53 & 76.17 & 0.466 & 0.110 & 18.98 & 86.35 & 0.584 & 0.091 \\
         & 1 & 11.68 & 74.95 & 0.445 & 0.110 & 18.26 & 86.32 & 0.568 & 0.092 \\
         & 2 & 11.41 & 74.97 & 0.435 & 0.109 & 17.68 & 86.03 & 0.553 & 0.093 \\
         & 3 & 11.00 & 74.67 & 0.424 & 0.108 & 16.97 & 85.81 & 0.536 & 0.094 \\
         & 4 & 10.87 & 75.03 & 0.419 & 0.107 & 16.46 & 85.67 & 0.521 & 0.096 \\
        \midrule
        \multirow{5}{*}{\textbf{Add-Global}} & 0 & 12.04 & 75.19 & 0.469 & 0.108 & 19.90 & 86.65 & 0.597 & 0.089 \\
         & 1 & 10.35 & 72.39 & 0.442 & 0.108 & 19.46 & 86.57 & 0.588 & 0.089 \\
         & 2 & 9.05 & 70.66	& 0.425 & 0.108 & 18.98 & 86.20 & 0.576 & 0.090 \\
         & 3 & 8.63 & 69.35 & 0.416 & 0.108 & 18.59 & 86.28 & 0.567 & 0.091 \\
         & 4 & 8.06 & 68.65 & 0.406 & 0.108 & 18.53 & 86.09 & 0.561 & 0.091 \\
        \bottomrule
    \end{tabular}
}
    \label{tab:sup_geal}
\end{table*}

\section{Corrupted Benchmark Construction} 
\label{sec:corrupted-benchmark}
To evaluate the robustness of language-prompted 3D affordance grounding models, GEAL~\cite{lu2024geal} introduced a set of corruption types applied to 3D point clouds. However, our work addresses intention-driven 3D affordance grounding, which requires both 3D point clouds and their corresponding reference images. Accordingly, we construct a corrupted benchmark by applying the same corruption types and severity levels as defined in GEAL to the 3D point clouds. The corruption types include scale, jitter, rotation, dropout-local, dropout-global, add-local, and add-global. Each type is evaluated under five severity levels, ranging from 0 to 4. For implementation details regarding the operation of these corruptions, we refer readers to the supplementary material of GEAL~\cite{lu2024geal}. Since the PIAD dataset provides multiple interaction images for each object, we randomly select one reference image from the available images for each corrupted 3D object. For every combination of corruption type and severity level, the benchmark contains 2,159 image-point cloud pairs, covering 23 object categories and 17 affordance types. The detailed statistics of the corrupted benchmark are summarized in~\tabref{tab:sup_geal_count}.

\section{More Visualizations} \label{sec:more-vis}
\subsection{Visualization on PIADv2}
\begin{figure*}[!htbp]
    \centering
    \includegraphics[width=0.99\linewidth]{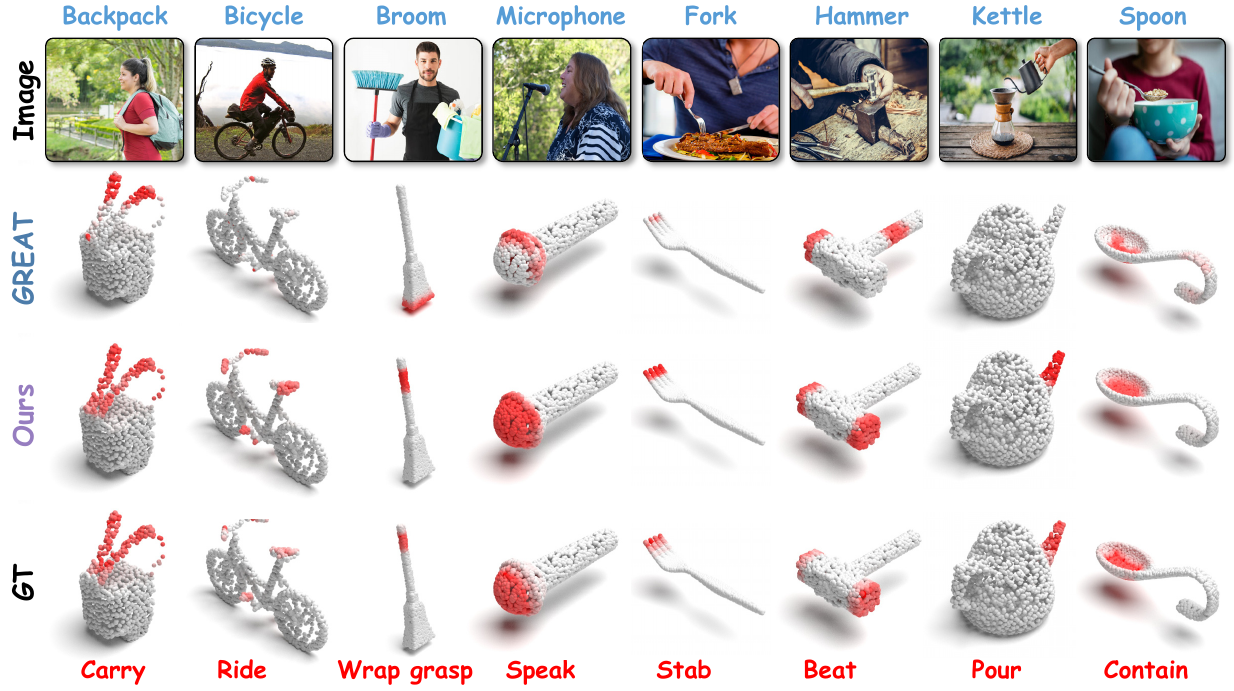}
    \caption{\textbf{Visualization on PIADv2 Seen.} We compare our~\ours with GREAT~\cite{shao2024great} on several examples from the Seen subset of PIADv2.}
    \label{fig:sup_piadv2_seen}
\end{figure*}

\begin{figure*}[!htbp]
    \centering
    \includegraphics[width=0.99\linewidth]{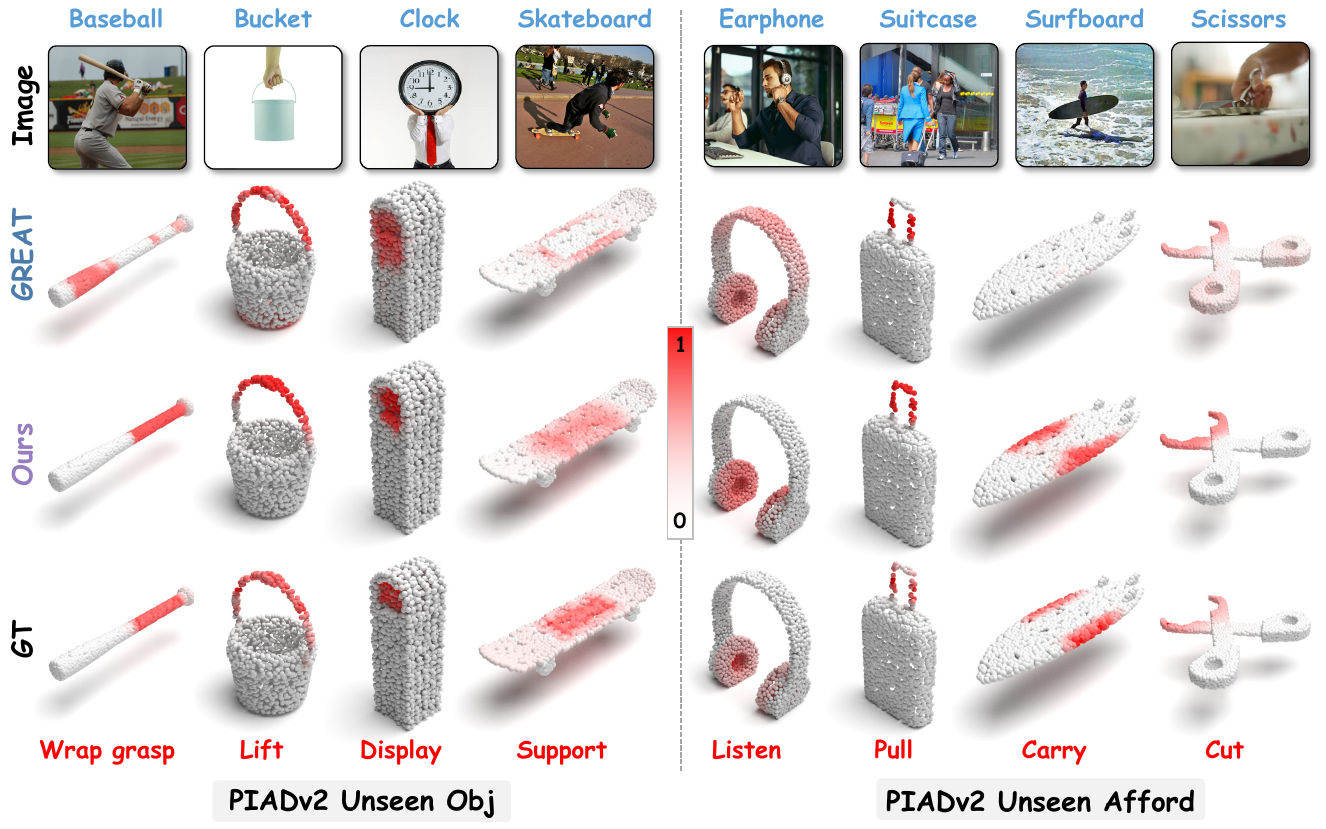}
    \caption{\textbf{Visualization on PIADv2 Unseen Object and Unseen Affordance subsets.} We compare our \ours~with GREAT~\cite{shao2024great} on several examples from the Unseen Object and Unseen Affordance subsets of PIADv2.}
    \label{fig:sup_piadv2_unseen}
\end{figure*}

To further illustrate the efficacy of our~\ours, we provide more qualitative comparisons on PIADv2~\cite{shao2024great}, complementing the results presented in the main manuscript. \figref{fig:sup_piadv2_seen} showcases additional visual examples on the Seen split of PIADv2, comparing our method with GREAT~\cite{shao2024great}. The results indicate that \ours consistently generates more accurate and complete affordance maps, which align more closely with the ground-truth annotations. For instance, in the case of the broom, our model correctly identifies the handle region suitable for the \textit{wrapgrasp} affordance, whereas GREAT erroneously highlights the cleaning area. Similarly, for the hammer, our approach precisely localizes the beating surface, while GREAT produces a less accurate region that includes portions of the handle. Furthermore, \figref{fig:sup_piadv2_unseen} provides more qualitative results on the Unseen Object and Unseen Affordance splits of PIADv2. These examples show the strong generalization capability of \ours when confronted with novel object categories and affordance types. For example, when presented with a bucket (an unseen object), our model accurately identifies the handle region for the \textit{lift} affordance, but GREAT yields an incomplete affordance map that omits critical parts. In another case involving the novel affordance \textit{listening} applied to earphones, our method successfully highlights the appropriate component, while GREAT's prediction includes irrelevant regions. These visual comparisons reinforce the effectiveness of \ours.

\subsection{Visualization of Point Features}
\begin{figure*}[!htbp]
    \centering
    \includegraphics[width=0.99\linewidth]{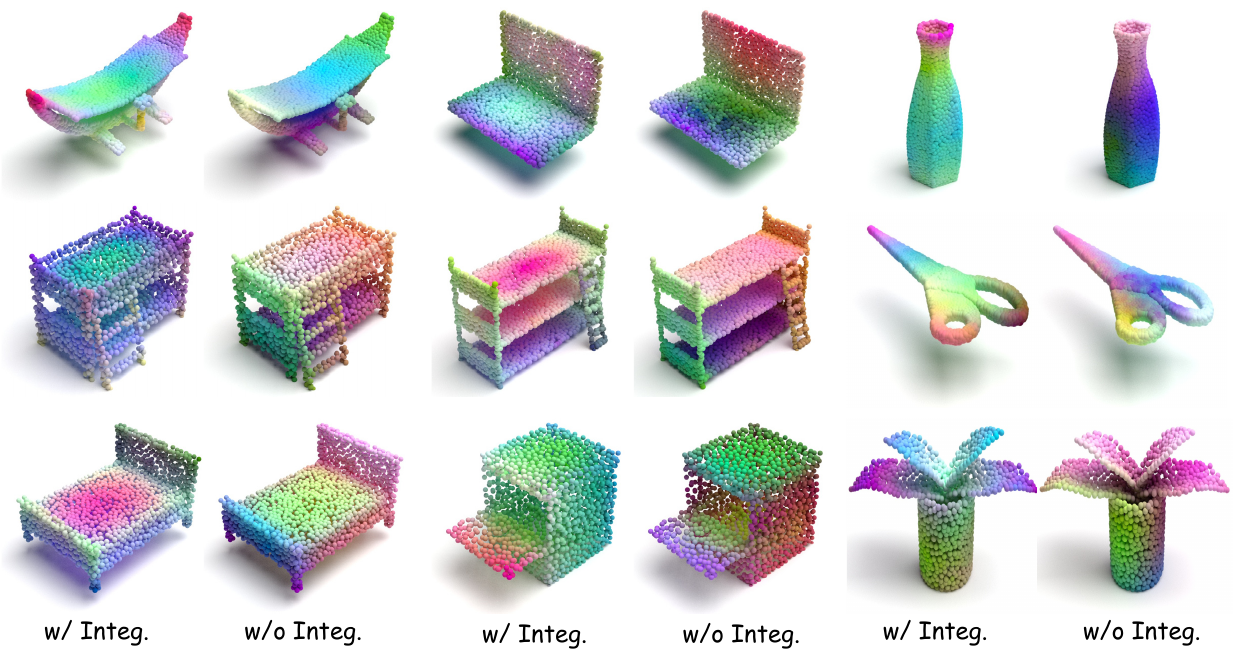}
    \caption{\textbf{Visualization of point features with and without hierarchical cross-modal integration.} We visualize the PCA-reduced point features for several examples to demonstrate the effectiveness of our proposed hierarchical cross-modal integration mechanism.}
    \label{fig:sup_integ}
\end{figure*}

In the main paper, we analyze the influence of our hierarchical cross-modal integration mechanism on point features by visualizing two examples of PCA-reduced features. This section provides additional visualizations in~\figref{fig:sup_integ} to offer a more comprehensive evaluation. The results clearly demonstrate that the proposed integration mechanism successfully incorporates contextual information from the reference image into the point features, yielding more discriminative and semantically enriched representations. For the bed shown in the second row, the point features are enhanced through our hierarchical integration form distinct clusters corresponding to key components such as the bed frame and other parts, which are essential for accurately grounding the \textit{lay} affordance. In contrast, the original features without integration appear more dispersed and less structured. Similarly, for the laptop and the bottle depicted in the first row, the integrated features exhibit improved separation between different functional regions, such as the keyboard and screen of the laptop, and the body and cap of the bottle. This enhancement in feature representation contributes to more precise affordance predictions, as supported by the qualitative results presented in the main text.

\begin{figure}[!htbp]
    \centering
    \includegraphics[width=0.85\linewidth]{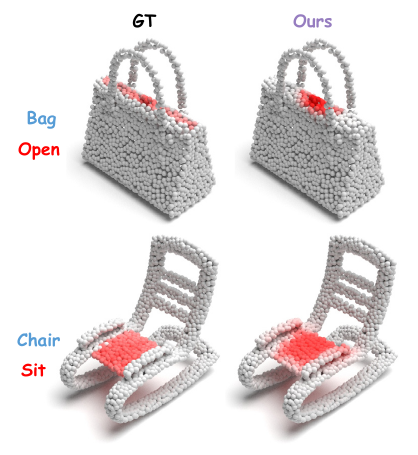}
    \caption{\textbf{Failure cases of our \ours~model on PIAD.}}
    \label{fig:sup_failure}
\end{figure}

\subsection{Visualization of Corrupted Point Clouds}
\begin{figure*}[!htbp]
    \centering
    \includegraphics[width=0.99\linewidth]{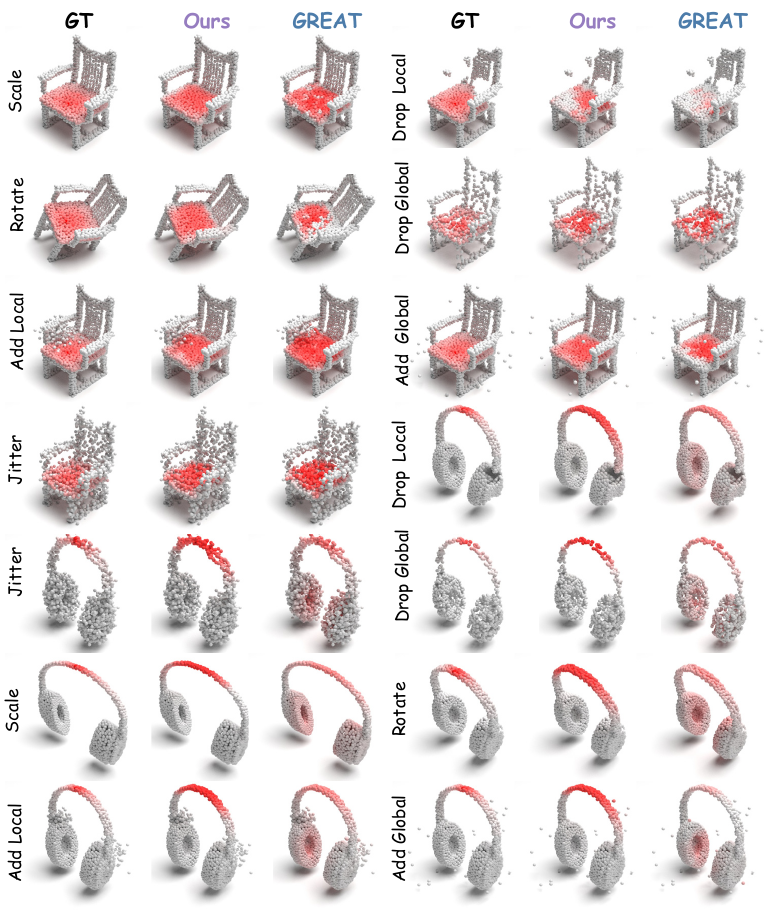}
    \caption{\textbf{Visualization of corrupted point clouds.} We illustrate two additional examples of corrupted point clouds from our constructed benchmark.}
    \label{fig:sup_geal}
\end{figure*}

Complementing the example in the main paper, we provide further qualitative results in~\figref{fig:sup_geal} to illustrate the corrupted point clouds utilized for robustness assessment. In the case of the \textit{sit} affordance on the chair, our \ours accurately identifies the seat region despite the presence of jitter and dropout corruptions, whereas GREAT~\cite{shao2024great} yields incomplete affordance maps. For the \textit{grasp} affordance on the earphone, \ours effectively localizes the handle area under rotation and additive noise corruptions, while GREAT's predictions are less precise and incorporate irrelevant regions. These additional visualizations show the robustness of \ours in handling diverse corruptions in 3D point clouds, enabling reliable affordance grounding in challenging scenarios.

\subsection{Failure Cases}
Although the proposed framework achieves competitive performance, it still exhibits certain limitations in challenging scenarios. \figref{fig:sup_failure} illustrates two representative failure cases. In the first example, which involves the \textit{open} affordance on a bag, the model fails to accurately localize the full zipper region suitable for opening and only activates in a limited area. In the second example, regarding the \textit{sit} affordance on a chair, the model incorrectly predicts the handrest region as part of the sittable area, indicating a misalignment between the inferred interaction and the object geometry. The error seems to arise from ambiguity in the intention embedding derived from the image, where the depicted interaction introduces uncertainty in distinguishing between relevant and irrelevant parts. These cases highlight specific challenges in disambiguating subtle visual cues and refining spatial awareness in complex interaction scenarios. Future work may explore more robust intention representation learning and enhanced geometric reasoning to overcome these limitations.

\clearpage

\end{document}